\ifdefined\useorstylema

\documentclass[opre,nonblindrev]{./macros/informs3}

\usepackage{./macros/packages1-or}
\usepackage{./macros/editing-macros}
\usepackage{./macros/statistics-macros-or}
\usepackage{./macros/formatting}

\DoubleSpacedXI % Made default 4/4/2014 at request
\usepackage{endnotes}
\let\footnote=\endnote

\usepackage{natbib}
 \bibpunct[, ]{(}{)}{,}{a}{}{,}%
 \def\bibfont{\small}%
 \def\bibsep{\smallskipamount}%
\TheoremsNumberedThrough     % Preferred (Theorem 1, Lemma 1, Theorem 2)
\ECRepeatTheorems

\EquationsNumberedThrough    % Default: (1), (2), ...
\usepackage{hyperref}

\usepackage{pgfplotstable}
\usepackage{graphicx}
\usepackage{subcaption}
\usepackage{float} 
\usepackage{tabularx}
\usepackage{overpic}
\usepackage{tikz}
\usepackage{rotating}
\usepackage{psfrag}
\usepackage{array}
  
\begin{document}
%%%%%%%%%%%%%%%%

%\setcounter{page}{1}
%\DOI{10.1287/xxxx.0000.0000}%


\RUNAUTHOR{}
\RUNTITLE{}

% \TITLE{Bundling Information Goods of Decreasing Value}
\TITLE{}

\ARTICLEAUTHORS{%

\AUTHOR{}
\AFF{
  \EMAIL{}} %, \URL{}}

} % end of the block

\ABSTRACT{For tabular datasets, the change in the relationship between the label and covariates ($Y|X$-shifts) is common due to missing variables (a.k.a. confounders). Since it is impossible to generalize to a completely new and unknown domain, we study models that are easy to adapt to the target domain even with few labeled examples. 
We focus on building more informative representations of tabular data that can mitigate $Y|X$-shifts, and propose to leverage the prior world knowledge in LLMs by serializing (write down) the tabular data to encode it. We find LLM embeddings alone provide inconsistent improvements in robustness, but models trained on them can be well adapted/finetuned to the target domain even using 32 labeled observations. Our finding is based on a comprehensive and systematic study consisting of 7650 source-target pairs and benchmark against \textbf{261,000} model configurations trained by 22 algorithms. Our observation holds when ablating the size of accessible target data and different adaptation strategies.
The code is available at \url{https://github.com/namkoong-lab/LLM-Tabular-Shifts}.}

%\KEYWORDS{deterministic inventory theory; infinite linear programming duality;
%  existence of optimal policies; semi-Markov decision process; cyclic schedule}

% \HISTORY{This paper was first submitted on July,
%   2020.}

\maketitle
%%%%%%%%%%%%%%%%%%%%%%%%%%%%%%%%%%%%%%%%%%%%%%%%%%%%%%%%%%%%%%%%%%%%%%

% Samples of sectioning (and labeling) in MNSC
% NOTE: (1) \section and \subsection do NOT end with a period
%       (2) \subsubsection and lower need end punctuation
%       (3) capitalization is as shown (title style).
%
%\section{Introduction.}\label{intro} %%1.
%\subsection{Duality and the Classical EOQ Problem.}\label{class-EOQ} %% 1.1.
%\subsection{Outline.}\label{outline1} %% 1.2.
%\subsubsection{Cyclic Schedules for the General Deterministic SMDP.}
%  \label{cyclic-schedules} %% 1.2.1
%\section{Problem Description.}\label{problemdescription} %% 2.

% Text of your paper here

\else

\documentclass[11pt]{article}
\usepackage[numbers]{natbib}

\usepackage{subfloat}
\usepackage{subfig}
\usepackage{./macros/packages}
\usepackage{./macros/editing-macros}
\usepackage{./macros/formatting}
\usepackage{./macros/llm-dro-macros}
\usepackage{./macros/statistics-macros}
\usepackage[toc,page]{appendix} 
\usepackage{minitoc}
\usepackage{tcolorbox}
 \usepackage{booktabs}
 \usepackage{multirow}
 \usepackage{graphicx}

\usepackage{lipsum}

\renewcommand{\thefootnote}{\fnsymbol{footnote}}

\begin{document}

% Control whitespace around equations
\abovedisplayskip=8pt plus0pt minus3pt
\belowdisplayskip=8pt plus0pt minus3pt

%\doparttoc % Tell to minitoc to generate a toc for the parts
%\faketableofcontents % Run a fake tableofcontents command for the partocs
%\thepart{} % Start the document part
%\parttoc % Insert the document TOC

% ------------------------------------------------------------------------
% Main Paper Body
% ------------------------------------------------------------------------

% ------------------------------------------------------------------------
% Default title and authorship
% ------------------------------------------------------------------------
\begin{center}
  %{\LARGE Index Policy for a Single-server Queue with Predicted Job Types } \\
 {\LARGE LLM Embeddings Improve Test-time Adaptation to \\
 \vspace{.1cm}
 Tabular $Y|X$-Shifts} \\ 
  \vspace{.5cm}
  {\Large Yibo Zeng\footnote{Equal Contributions}$^{,1}$, Jiashuo Liu$^{*,2}$, Henry Lam$^1$, Hongseok Namkoong$^1$} \\ 
  \vspace{.2cm} 
  {\large Columbia University$^1$, Tsinghua University$^2$}  \\
  \vspace{.2cm}
  \texttt{yibo.zeng@columbia.edu}
  \hspace{1cm}
  \texttt{liujiashuo77@gmail.com}\\
    
    \texttt{khl2114@columbia.edu}
    \hspace{1cm}
    \texttt{namkoong@gsb.columbia.edu}
\end{center}

% ------------------------------------------------------------------------
% Abstract
% ------------------------------------------------------------------------

\begin{abstract}%
For tabular datasets, the change in the relationship between the label and covariates ($Y|X$-shifts) is common due to missing variables (a.k.a. confounders). Since it is impossible to generalize to a completely new and unknown domain, we study models that are easy to adapt to the target domain even with few labeled examples. 
We focus on building more informative representations of tabular data that can mitigate $Y|X$-shifts, and propose to leverage the prior world knowledge in LLMs by serializing (write down) the tabular data to encode it. We find LLM embeddings alone provide inconsistent improvements in robustness, but models trained on them can be well adapted/finetuned to the target domain even using 32 labeled observations. Our finding is based on a comprehensive and systematic study consisting of 7650 source-target pairs and benchmark against \textbf{261,000} model configurations trained by 22 algorithms. Our observation holds when ablating the size of accessible target data and different adaptation strategies.
The code is available at \url{https://github.com/namkoong-lab/LLM-Tabular-Shifts}.
\end{abstract}

\fi 

\section{Introduction} \label{section: introduction}

Predictive performance degrades when the distribution of target domain shifts from that of  source (training)~\citep{BandiEtAl18,WongEtAl21, Hand06,DingHaMoSc21, AmorimCaVe18}. 
Distribution shifts can be categorized into shifts in the marginal
distribution of covariates ($X$-shifts) or changes in the  relationship between the label and covariates ($Y|X$-shifts). 
In computer vision, $X$-shifts are prevalent since high-quality human labels are consistent across different images~\citep{RechtRoScSh19,MillerTaRaSaKoShLiCaSc21, ShankarDaRoRaReSc19};
in contrast, $Y|X$-shifts are prevalent in tabular data due to missing variables and hidden confounders. 
There is a large body of work addressing $X$-shifts
due to its dominance in vision and language~\citep{li2017deeper,zhuang2020comprehensive, zhou2022domain}, yet the work on $Y|X$-shifts remain relatively limited~\citep{LiuWaCuNa23}.

The main challenge with addressing $Y|X$-shifts in tabular tasks
is that the source data may provide little insight on the target distribution. 
Since it is impossible to generalize to a completely new and unknown domain~\citep{ArjovskyBoGuLo19,RosenfeldRaRi20}, we focus on leveraging few labeled target examples (on the order of 10 to 100) to address small $Y|X$-shifts
that negatively impact model performance. 
Our goal is to build a feature representation $\phi(X)$ such that
the difference between
$\E_{\rm source}[Y|\phi(X)]$ and $\E_{\rm target}[Y|\phi(X)]$ 
are learnable even based on a few target data.

\begin{figure}[t]
\includegraphics[width=1\textwidth]{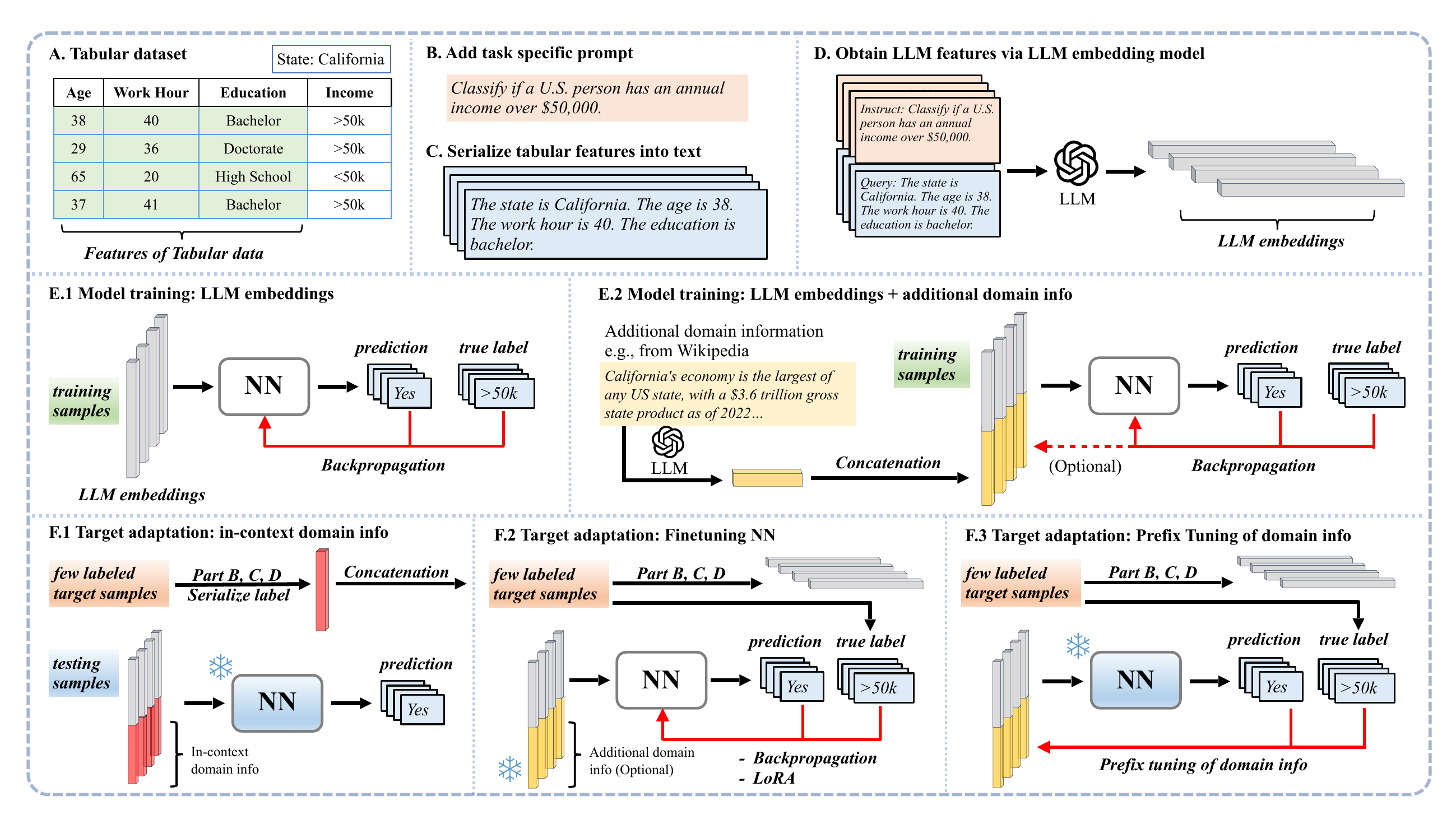}
\caption{Overview of methods incorporating LLM embeddings.}
\label{fig: method overview}
\end{figure}

Using the wealth of world knowledge learned during pre-training, LLMs have the potential to build representations 
that mitigate the impact of confounders whose distribution changes across source and target.
Specifically, we use a recent and advanced LLM encoder (\texttt{e5-Mistral-7B-Instruct}~\cite{WangEtAl23}) to featurize tabular data---which we referred to as LLM embeddings---and fit a  shallow neural network (NN) on these embeddings for tabular prediction (\Cref{fig: method overview}).
In contrast to classical numerical encoding of tabular data,
our approach automatically incorporates the semantics of each covariate using off-the-shelf LLMs, and can include additional contextual domain-level information that can help account for missing variables whose distribution shifts from source to target. 

Throughout our investigation, we use the same LLM encoder to extract the LLM embeddings, and we only ``finetune'' the shallow NNs we use as the main prediction model across target domains. Investigating how different LLM encoders, e.g., LLMs specialized for tabular data~\citep{YanEtAl24}, affect tabular $Y|X$ shifts is left as a future work.
Comparison to other LLM-based tabular classification is discussed in related work below.

For rigorous empirical evaluation, we consider \textbf{7,650} natural spatial shifts (source$\rightarrow$target) based on three real-world tabular datasets (\texttt{ACS Income, Mobility, Pub.Cov}~\citep{DingHaMoSc21}). Our testbed serves as a  \emph{large-scale} benchmark for $Y|X$-shifts on tabular data, offering a standardized protocol for training, validation, testing, and finetuning, as well as a consistent hyperparameter selection process.  
Compared to previous benchmarks on tabular distribution shifts~\citep{LiuWaCuNa23}, this paper not only explores a significantly greater variety of shift settings but also introduces a series of novel approaches to incorporate LLM embeddings as features.
Such a comprehensive evaluation ensures the robustness and adaptability of our findings across diverse scenarios, setting a new standard for future work in this domain.
We compare our proposed approach with: (i) typical methods on \textsf{Tabular} features, including basic models (LR, SVM, NN), gradient-boosting trees (GBDT; XGB, LGBM, GBM), distributionally robust methods (DRO; KL-DRO, $\chi^2$-DRO, Wasserstein DRO, CVaR-DRO, and Unified-DRO); and (ii) in-context learning~\cite{BrownEtAl20} using recent and advanced pre-trained LLMs, including TabPFN~\cite{HollmannMuEgHu22} and GPT-4o mini~\cite{OpenAI24}. See~\Cref{tab:overview-methods} for further details.
 In total, we consider 22 algorithms and  \textbf{261,000} model configurations.

\begin{figure}[t]
  \centering
  \subfloat[\texttt{ACS Income}]{\includegraphics[width=0.25\textwidth]{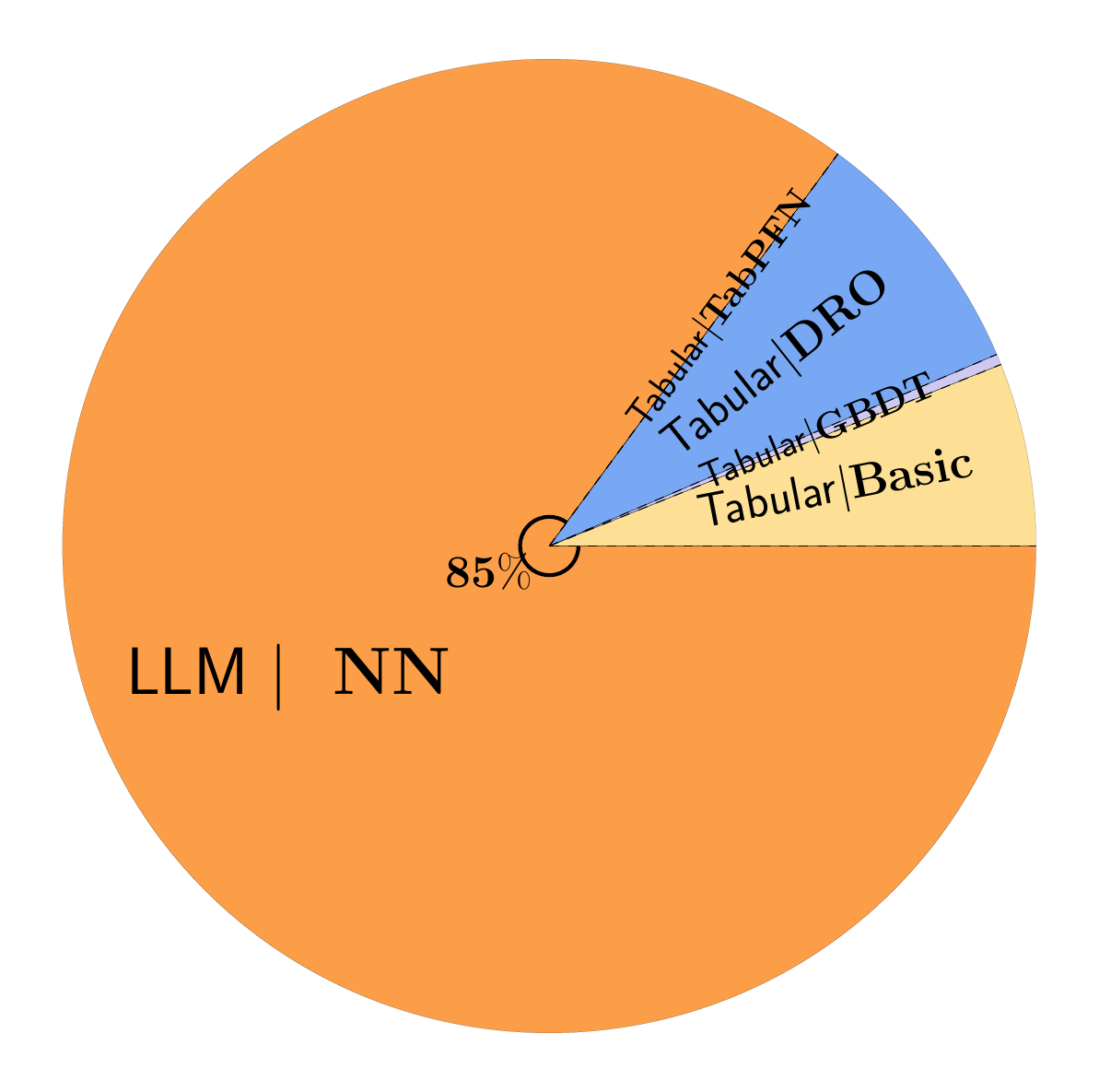}
  \label{fig6:subfig1}}
  \subfloat[\texttt{ACS Mobility}]{\includegraphics[width=0.25\textwidth]{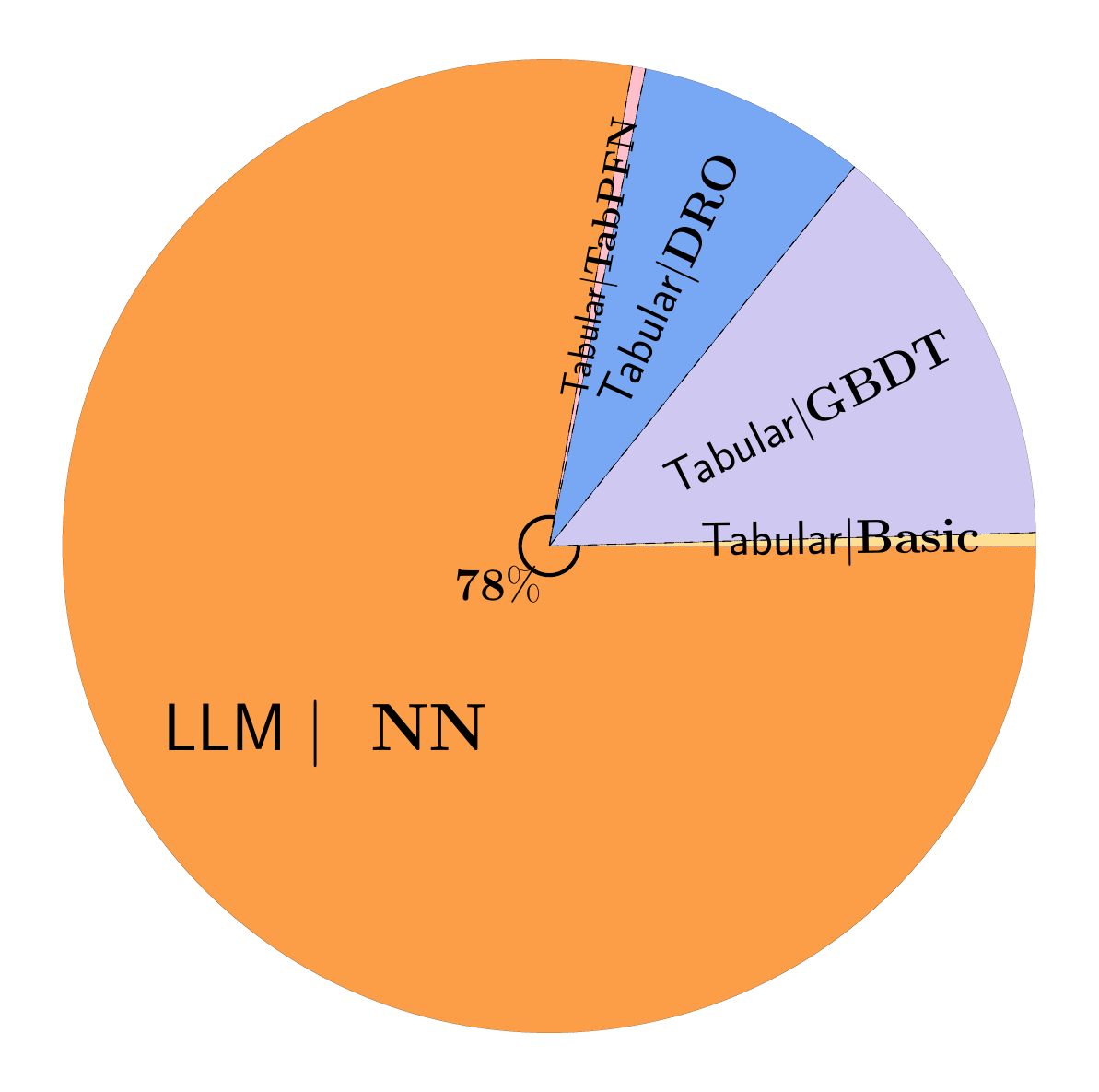}\label{fig6:subfig2}}
  \subfloat[\texttt{ACS Pub.Cov}]{\includegraphics[width=0.25\textwidth]{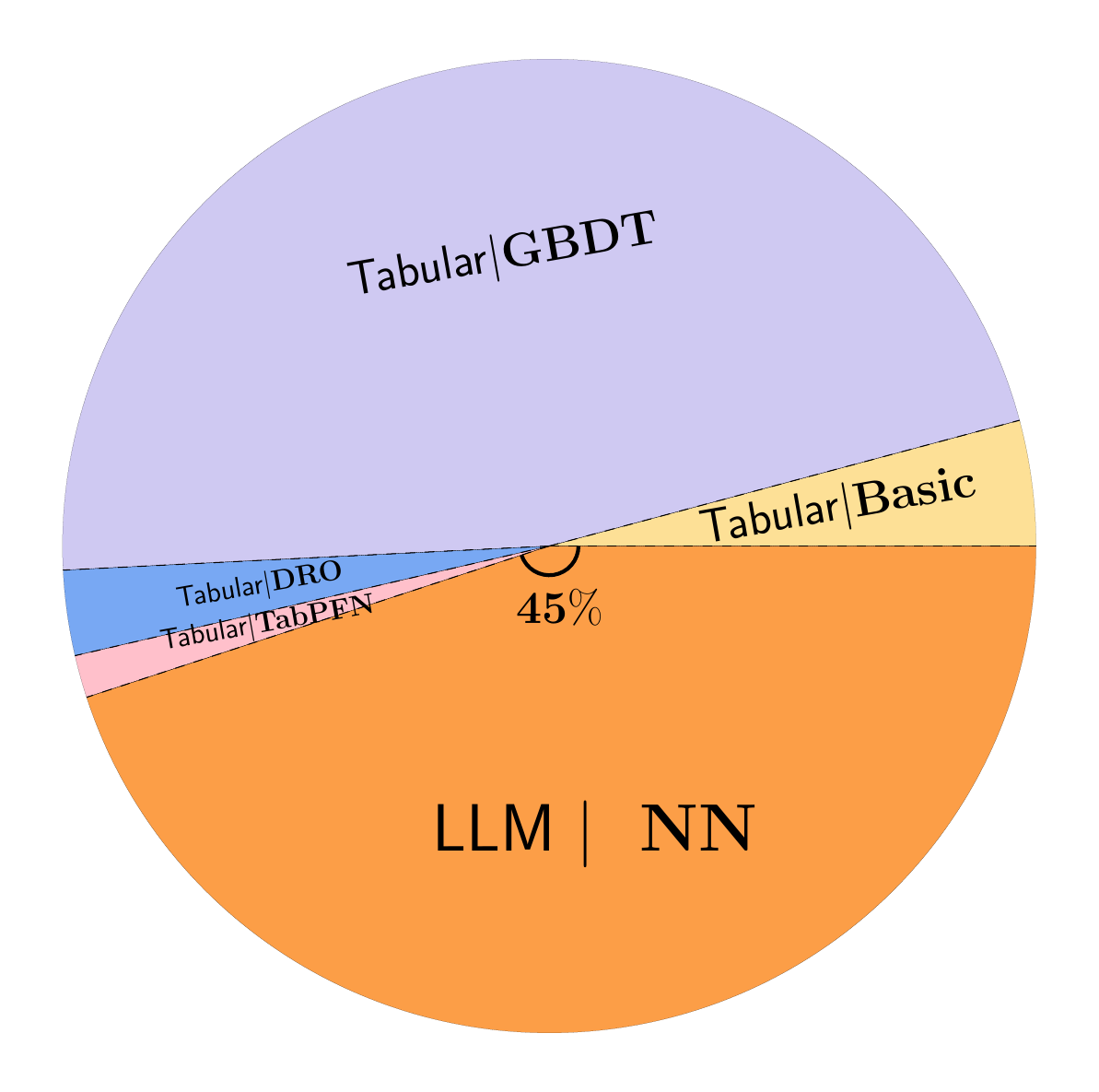}\label{fig6:subfig4}}\\
    \subfloat[\texttt{ACS Income}]{\includegraphics[width=0.25\textwidth]{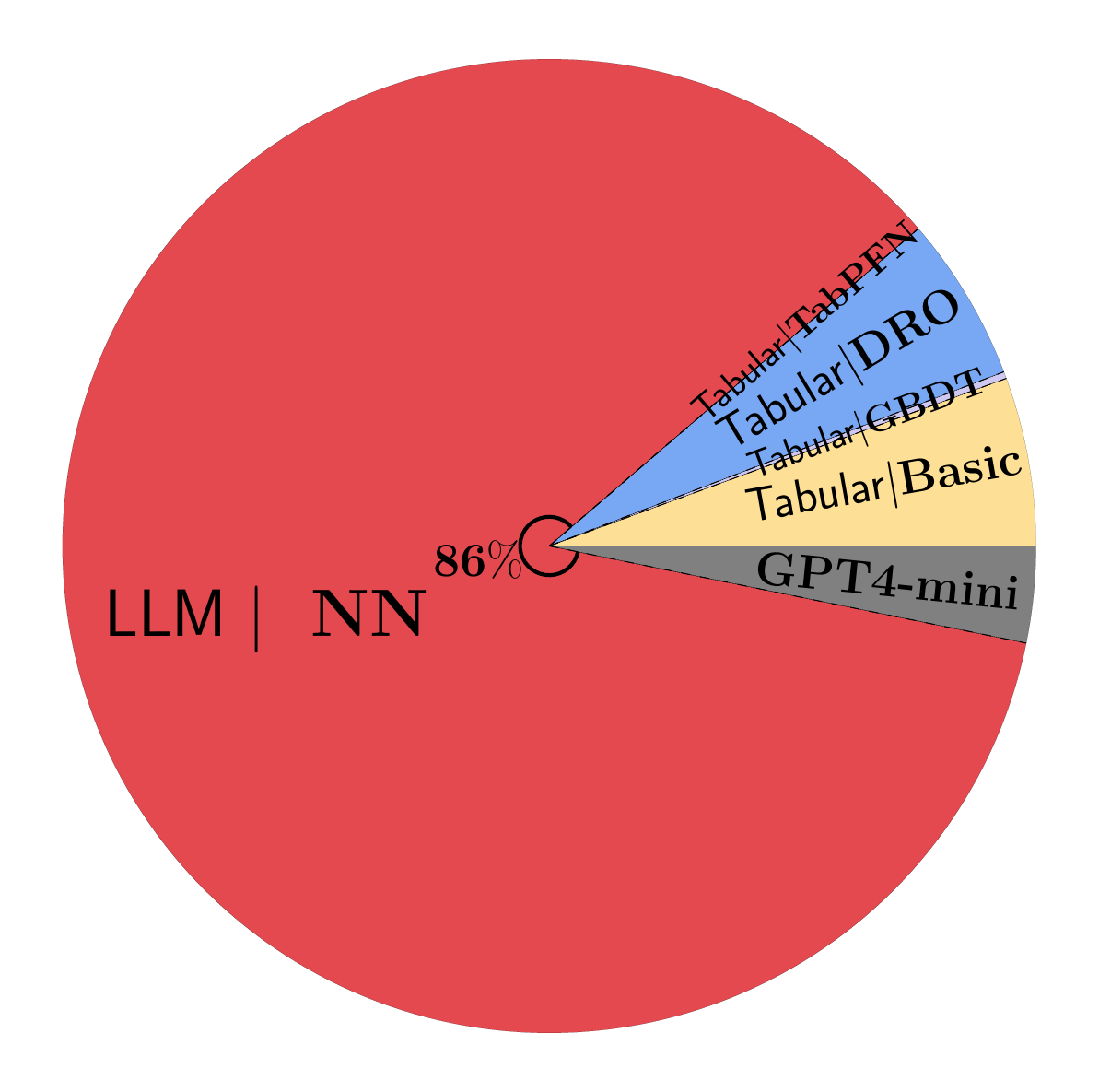}
  \label{fig6:subfig4}}
  \subfloat[\texttt{ACS Mobility}]{\includegraphics[width=0.25\textwidth]{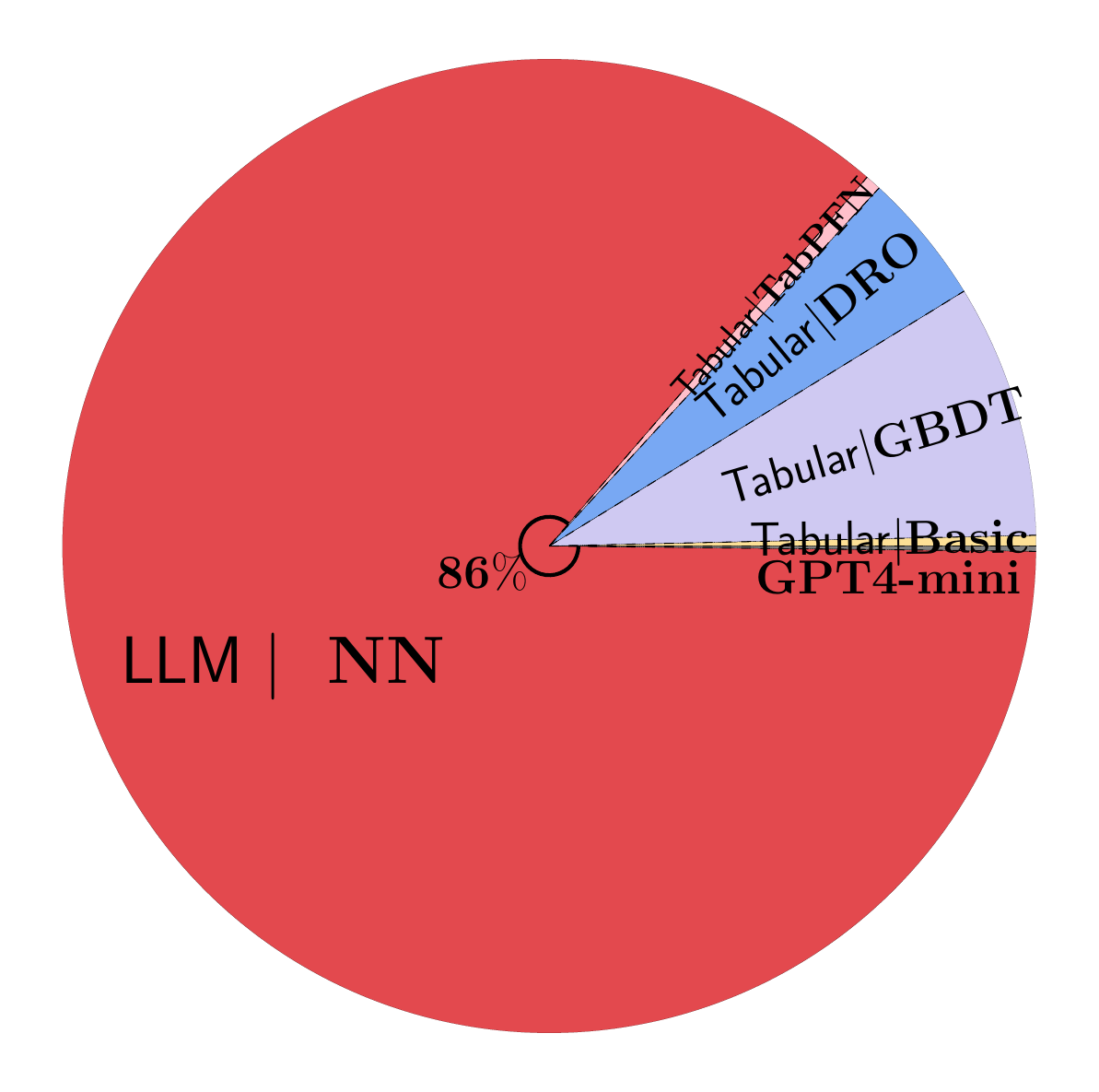}\label{fig6:subfig5}}
  \subfloat[\texttt{ACS Pub.Cov}]{\includegraphics[width=0.25\textwidth]{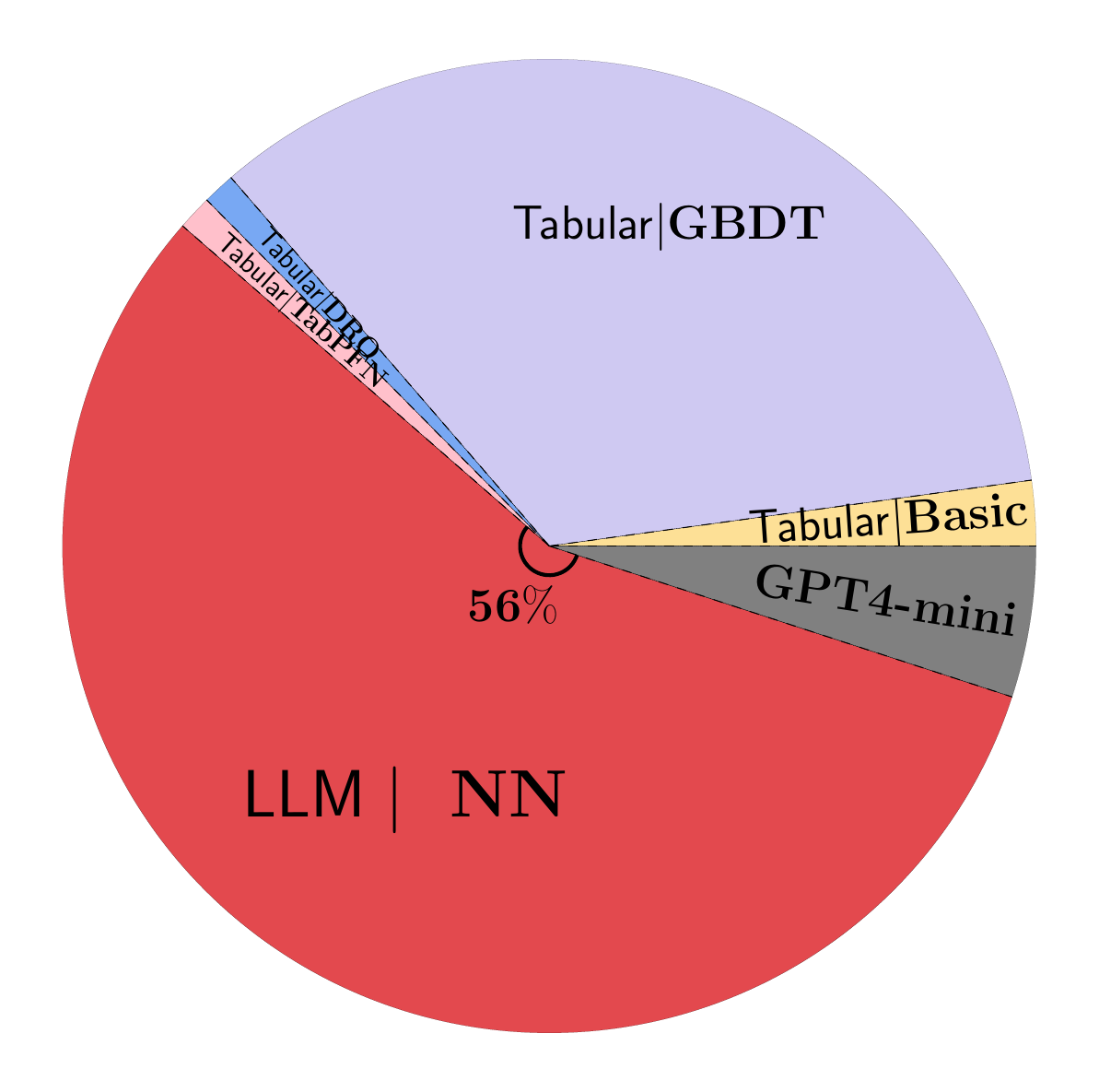}\label{fig6:subfig6}}
  \caption{The FractionBest Ratio in~\Cref{equ:optimal-ratio} (with $\Delta=1\%$). We compare our proposed methods---(a)-(c): \textsf{LLM}$|$NN and (d)-(f): \textsf{LLM}$|$NN (finetuning)---with methods on \textsf{Tabular} features.}
 \label{fig:optimal-ratio}  
\end{figure}

Since it is unrealistic to expect any single method to uniformly dominate over large number of source$\rightarrow$target settings, we complement traditional average-case metrics using the \emph{fraction of times each method performs best}. For each method $\mathcal{M}$, 
\begin{equation}
\label{equ:optimal-ratio}
	\text{FractionBest}(\mathcal M; \Delta) := \frac{|\mathcal S(\Delta)\cap \mathcal S_{\mathcal M}|}{|\mathcal S(\Delta)| },
\end{equation}
where $\mathcal S(\Delta)$ contains all source$\rightarrow$target settings where the performance between the best and second best model is larger than $\Delta$ (we set $\Delta=1\%$ in this paper), and $\mathcal S_{\mathcal M}$ contains all source$\rightarrow$target settings where model $\mathcal M$ performs the best.
$\text{FractionBest}$ calculates the proportion of source→target settings where (i) $\mathcal{M}$ outperforms all other methods and (ii) the improvement over the second-best model is meaningful (and significant).  

First, we consider LLM embeddings without any adaptation to labeled target data\footnote{We do not conduct any target adaptation; however,  we use 32 labeled target samples for validation (selection of hyperparameters, etc.).}. Shallow networks based on LLM embeddings (\textsf{LLM}$|$NN) outperform all other methods on tabular features in \textbf{85\%} settings in the \texttt{ACS Income} dataset, and in \textbf{78\%} in the \texttt{ACS Mobility} dataset. However, for the \texttt{ACS Pub.Cov} dataset, the FractionBest drops to \textbf{45\%}, which indicates that LLM embeddings do not always offer a perfect solution (see \Cref{fig:optimal-ratio} (a)-(c)). We conclude \emph{LLM embeddings sometimes improve robustness}, but \emph{do not consistently surpass} state-of-the-art tree-ensemble methods.

However, we find that \emph{finetuning} the prediction model (shallow NN)
 \emph{with few target samples} can \emph{make a big difference} even when using identical LLM embeddings. When finetuning with just 32 target samples for finetuning, the FractionBest ratio (\Cref{equ:optimal-ratio}) improves from 85\% to \textbf{86\%}  on \texttt{ACS Income}, from 78\% to \textbf{86\%} on \texttt{ACS Mobility},  and from 45\% to \textbf{56\%} on \texttt{ACS Pub.Cov} (see \Cref{fig:optimal-ratio} (d)-(f)).
We find this improvement \emph{surprising}:
although the shallow NN has numerous parameters, finetuning with only 32 target samples surprisingly improve target performance by a relatively large margin. More importantly, such improvement is observed under $Y|X$ shifts and  holds across a wide range of distributional shift settings. 
Furthermore, our approach significantly outperforms in-context learning using advanced pre-trained models like TabPFN~\cite{HollmannMuEgHu22} and GPT-4o mini~\cite{OpenAI24}, while being much more lightweight.  We only need to finetune shallow NNs, rather than engage in extensive engineering efforts required for in-context learning, such as prompt engineering, or the costly end-to-end updates involved in few-shot fine-tuning of LLMs (see further comparison in related work below).
This suggests a promising new research direction for further investigation of using LLM embeddings in tabular classification tasks. 
Theoretical insights are discussed in Section~\ref{sec: theory}.

Our method also implies multiple additional benefits (See Section~\ref{sec: primary finding}). 
%With the same amount of target samples, the finetuned NNs significantly outperform in-context learning with GPT-4o mini, which can be viewed as the SOTA decoder model (see Figure~\ref{fig:optimal-ratio} (d)-(f)). 
\emph{The performance gain brought by target samples is larger under stronger $Y|X$-shifts}, where the level of $Y|X$-shifts is measured by  DISDE~\citep{CaiNaYa23}. Finetuning with 32 target samples yields an average performance gain of 5.4 percentage points across the worst 500 settings on \texttt{ACS Pub.Cov}, compared to no finetuning. This is notably higher than the 1.2\% average gain observed across all 2550 settings (\textbf{4.5 times}).

Beyond our primary findings, we also conduct ablation studies to better understand our approach in Section~\ref{sec: aux finding}. 
First, incorporating the \emph{``right'' domain information has an outsize impact} on tabular $Y|X$-shifts. As shown in~\Cref{fig:curve} (f), for \texttt{ACS Pub.Cov} with finetuning,  adding additional domain information from Wikipedia improve F1 scores by 1.4pp on average across all 2550 settings and  1.7pp across the worst 500 settings when compare to using LLM embeddings alone. However, identifying the best domain information requires non-trivial engineering effort, which we left as a future work. Moreover, 
given the large number of model parameters and limited labeled target samples, one might expect parameter-efficient methods like Low-Rank Adaptation (LoRA)~\citep{hulora} and Prefix Tuning~\citep{li2021prefix} to offer a clear advantage for target adaptation. However, we find \emph{the specific finetuning approach has small impact} on target adaptability under tabular $Y|X$-shifts.
In~\Cref{fig:lora},  all target adaptation methods significantly outperform the non-finetuned version, both when using LLM embeddings alone and using LLM embeddings with domain information from Wikipedia.

Another practical question is how to allocate a fixed number of labeled target samples between target adaptation and validation (selection of finetuning method, hyperparameters, etc). 
In Figure~\ref{fig:allocation} to come,
we compare two allocation schemes of $64$ labeled target samples: (i) using all 64 samples for validation (solid bar), and (ii) dividing the samples into 32 for validation and 32 for finetuning (shaded bar). For \texttt{ACS Mobility} and \texttt{ACS Pub.Cov}, \emph{target adaptation provides significant gains over the validation-only approach}, highlighting the need for further investigation
into sample allocation.

\vspace{-0.1in}
\paragraph{Related work}
Tabular data is a common modality in electronic health records, finance, and social and natural sciences.
Unlike other modalities like images and text, 
gradient-boosted trees (GBDT; GBM~\citep{friedman2001greedy,friedman2002stochastic}, 
XGBoost~\citep{chen2016xgboost}, LGBM~\citep{ke2017lightgbm}) remain the
state-of-the-art~\citep{gorishniy2021revisiting, shwartz2022tabular,gardner2022subgroup}
even when compared to 
neural networks specifically designed for tabular data~\citep{arik2021tabnet,huang2020tabtransformer,kadra2021well,katzir2020net}. 
GBDTs have recently been observed to provide strong performance under distribution shifts, which forms the basis of its use as a main baseline~\citep{gardner2022subgroup, LiuWaCuNa23}. 
We train GBDTs using only tabular features and do NOT use LLM embeddings because GBDTs are known to be slow to train on high-dimensional inputs~\cite{ke2017lightgbm}.

Recently, there exists another line of work that uses LLMs for tabular data classification. Unlike our approach that performs \emph{lightweight} probing using an encoder LLM, most work use a decoder LLM to pose the classification task as a question and predict the next token (e.g., ``Yes" for the positive class and ``No" for the negative class~\cite{hegselmann2023tabllm, SlackSi23}). These work can be categorized into three groups: (1) few-shot finetuning that uses labeled target to finetune or lightweight finetune the LLM~\cite{hegselmann2023tabllm, WangWaSu23, DinhEtAl22}; (2) in-context learning~\cite{WenEtAl24, SlackSi23, HollmannMuEgHu22} that uses few labeled target data as the prompt; (3) zero-shot learning~\cite{hegselmann2023tabllm, SlackSi23}, which assumes no access to the target data. The decoder can be a general LLM (e.g., T0 as in~\cite{hegselmann2023tabllm}) or finetuned on general tabular data~\cite{HollmannMuEgHu22}. 
In this paper, due to the large-scale testbed we consider, we do NOT consider expensive end-to-end updates~\citep{hegselmann2023tabllm, yang2024unleashing} to LLMs using source or target data, which requires full weight access and weight updates to the LLM. This approach is more computational expensive than finetuning a shallow NN, and  we find it infeasible across thousands of source-target pairs we consider. 
We leave few-shot finetuning as potential future work. 

More importantly, existing LLM work for tabular  classification  does not explicitly address distributional shifts. In particular, zero-shot learning struggles to generalize across target domains with varying levels of $Y|X$ shifts, making it impractical in our setting.  Moreover, while in-context learning adapts to labeled target samples, it generally  cannot utilize the large amount of labeled source data due to token limits\footnote{An alternative can be finetuning the decoder on source data and then apply in-context learning with labeled target samples. However, to our best knowledge, this has NOT been explored in the literature. Also, end-to-end updates to the LLM are costly and impractical for our large-scale testbed.}. Since the $Y|X$ distribution in the source domain may be relevant to the target domain, this limits the suitability of in-context learning for tabular $Y|X$ shifts.
As shown in Figure~\ref{fig:optimal-ratio}, we consider two baselines for this line of research: (i) TabPFN\footnote{TabPFN~\cite{HollmannMuEgHu22} can handle in-context learning with a relatively large number of samples, up to the level of 1000. Therefore, we use it for in-context learning on the source data, even though this is much smaller than the 20,000 source samples available.} with 1024 labeled source data~\cite{HollmannMuEgHu22}, which shows strong performance for tabular data classification with no distribution shift and outperform 18 other algorithms across 176 dataset in~\citep{mcelfresh2024neural}, and (ii) in-context learning using the  recent, advanced decoder GPT-4o mini and 32 labeled target sample.

A wide range of methods have been proposed to address distribution shifts, notably robust learning methods, balancing methods, and invariant learning methods. 
Distributionally robust optimization (DRO) construct an uncertainty set around the training distribution and optimizing for the worst-case distribution within this set, thereby mitigating the impact of potential distribution shifts.
Variants of DRO methods have been developed using different distance metrics, such as $\chi^2$-divergence~\citep{DuchiNa21,DuchiGlNa21}, KL-divergence~\citep{hu2013kullback}, and Wasserstein distance~\citep{BlanchetKaZhMu17, BlanchetMuZh18, BlanchetKaMu19, BlanchetKuLiTRa23}.
However, these approaches have recently been observed to be ineffective in addressing real-world tabular distribution shifts~\citep{LiuWaCuNa23}.
On the other hand, invariant learning~\citep{peters2016causal,ArjovskyBoGuLo19,koyama2020invariance} seeks to learn causally invariant relationships across multiple \emph{pre-defined} environments. In this work, we include 5 typical DRO methods but do not consider invariant learning methods as they require multiple training environments, which is not the focus of this work.
Some statistical works~\citep{li2022transfer, tian2023transfer} provide theoretical guarantees for simple linear models in transfer learning, but these guarantees often do not extend to more complex models like decision trees or neural networks used in real-world applications.
This version improves clarity and flow while retaining the original meaning.Additionally, many studies have explored domain adaptation~\citep{NEURIPS2021_1415fe9f, liang2023ttasurvey, chen2023improved}.
However, most of these focus on $X$-shifts in image data, whereas our work addresses $Y|X$-shifts in tabular data.

\section{Methods}
\label{sec:methods}

In this section, we introduce a series of methods utilizing LLM embeddings for tabular prediction, as well as different choices to incorporate additional domain information, different model architectures, and target adaptation techniques 
using a small amount of labeled target samples.
To the best of our knowledge, this work is the first to comprehensively explore the impact of LLM embeddings on tabular $Y|X$-shifts.

\subsection{LLM Embeddings for Tabular Prediction}
\label{section: llm embedding}

We first introduce how we transform tabular data into LLM embeddings, where the
key idea is to serialize each sample into a natural language format that the 
LLM can process.  
There is a substantial body of research on serialization, including using another LLM to rewrite tabular data into natural language~\citep{hegselmann2023tabllm}, adding descriptions of the classification task, training and test examples~\citep{hegselmann2023tabllm, SlackSi23}, etc. 
Among these methods, \citet{hegselmann2023tabllm} demonstrate that using 
a straightforward text template with a task description consistently 
achieves the best empirical performance.

Using an income prediction problem to illustrate, consider a simple task description such as ``\texttt{Classify whether US working adults' yearly income is above \$50000 in 2018.}'' along with a simple serialization template that enumerates all features in the format 
``\texttt{The [feature name] is [value]}''. 
Adopting this serialization approach, we employ the encoder model \texttt{e5-Mistral}-\texttt{7B-Instruct}~\cite{WangEtAl23} to generate the LLM embedding. Formally, the encoder takes the serialization $\text{Serialize($X$)}$ of sample $X$ as input and outputs its corresponding embedding $\Phi(X)$ as
\begin{equation*}
X \xrightarrow{\text{serialization}}\text{Serialize}(X) \xrightarrow{\texttt{e5-Mistral-7B-Instruct}} \Phi(X).
\end{equation*}
Since \texttt{e5-mistral-7b-instruct} requires input data to be formatted in the following template:
\begin{align*}
&\texttt{Instruct:} 
&&\texttt{description of the classification task \textbackslash n} \\
&\texttt{Query:} &&\texttt{description of the data},
\end{align*}
we provide task description in the ``Instruct" part, and use the serilization template to format the tabular data in the ``Query" part. 
An illustrative example is provided in Part A-D of~\Cref{fig: method overview},
with additional details available in 
Appendix~\ref{sec: appendix description of the data}.
Analyzing the impact of different LLM encoders, task descriptions, and serialization methods is left for future work.

\subsection{Additional Domain Information}
\label{section: additional domain info}
Another advantage of using LLM embeddings is their ability to incorporate additional domain information or prior knowledge, denoted by $C$. 
As demonstrated in Section~\ref{section: introduction}, incorporating domain-specific information can help address $Y|X$-shifts and improve generalization performance in the target domain.

In this work, we propose a simple yet effective approach for integrating domain knowledge into tabular predictions. 
Rather than combining the domain information with serialized tabular data and generating a single LLM embedding, we generate separate LLM embeddings for the domain knowledge and the serialized tabular data, and then concatenate them together.
The benefits of this approach are twofold: (a) although the domain information may contain significantly more words than the serialized tabular features, our concatenation method ensures a balanced 1:1 ratio between the two, preventing a single embedding that disproportionately focuses on the longer domain information; (b) by separating the tabular features from the domain information, we can efficiently update the domain information without having to regenerate all the embeddings for the entire dataset.

We explore three sources of domain information: Wikipedia, GPT-4~\cite{OpenAI23}, and labeled target samples. 
Given that our experiments (see \Cref{table:overview} and \Cref{section: experiments}) focus primarily on socioeconomic factors, we collected ``Economy" data for each U.S. state from Wikipedia as $C$. 
For GPT-4, we prompt it to provide background knowledge relevant to each prediction task in each state as $C$.
For labeled target samples, we serialize 32 labeled samples from the concerned domain as the prior knowledge $C$.
Further details can be found at Appendix~\ref{sec: appendix additional domain info}.
After obtaining domain information $C$, we use \texttt{e5-mistral-7b-instruct} to generate an LLM embedding for $C$. 
As illustrated in Parts E.2 and F.1 of Figure~\ref{fig: method overview}, this embedding is then concatenated with the LLM embeddings of the tabular data, which serve as input to the backend neural network models (NN). 
This approach allows us to generate the LLM embedding for the dataset \emph{just once}, and subsequently concatenate it with embeddings from different prompts as needed.
In Section~\ref{section: experiments}, we study whether and how this 
additional domain information can enhance generalization under $Y|X$-shifts.

In addition, recent works on prompt engineering have focused on incorporating additional domain information to enhance prediction tasks, often through detailed instructions~\citep{SchickSc20, ShinEtAl20}. 
Our proposed framework introduces a novel approach to leveraging such information and remains fully compatible with these existing methods.

\subsection{Model Training and Target Adaptation}
\label{section: model training and adaptation}

\vspace{-0.02in}
\paragraph{Model architecture}
For the backend model, we use a vanilla neural network (NN) classifier on both tabular features and LLM embeddings for tabular data classification. 
The NN is a simple feedforward neural network with several hidden layers, dropout layer, and ReLU activation functions. 

When adding additional domain information via an embedding layer, the same embedding is applied to all samples from the same domain.
Since the output of \texttt{e5-mistral-7b-instruct} is a 4096-dimensional vector, 
we simply concatenate the LLM embeddings with the embeddings of the domain information. This concatenated vector is then 
passed through the hidden layers, dropout layer, and ReLU activation functions. For all NNs, the final linear layer an output dimension of 2, followed by a softmax layer for binary classification. 
During training, we use cross-entropy as the loss function, batch size as 128, 
and use the Adam optimizer.  
Detailed model architecture and hyper-parameters are provided in the 
Appendix~\ref{sec:appendix model arch} and~\ref{sec: appendix hps},
with a discussion on hyperparameter selection provided in 
Section~\ref{sec: testbed setup}.

\vspace{-0.1in}
\paragraph{Target Adaptation}
Even with the incorporation of LLM embeddings and domain information, our model may still experience $Y|X$-shifts. 
In practice, it is common to have a small set of samples from the target domain, which can be leveraged to better adapt the model to the target domain. 

For each (source domain, target domain) pair, we begin by selecting the best 
training hyperparameter based on a validation criterion, which will be 
discussed in the Section~\ref{sec: testbed setup}. 
Using this model trained on the source domain,  we explore four primary
methods for target adaptation: in-context domain info, full-parameter fine-tuning, 
low-rank adaptation (LoRA), and prefix tuning for domain information. 

For in-context domain info (F.1 of Figure~\ref{fig: method overview}), 
we keep the trained model frozen and only update the 
domain information, switching it from natural language description of labeled sample from the source domain during training to that of target domain during inference phase.  
For the other three methods, we conduct further training of the model. In full-parameter fine-tuning (F.2 of \Cref{fig: method overview}), the entire neural network is fine-tuned using the target samples. For LoRA, 
we introduce a low-rank adaptation layer to each linear layer by incorporating two smaller matrices, $A$ and $B$, both with a rank of 16. Specifically, matrix $A$ has dimensions corresponding to the input size and the rank, while matrix $B$ has dimensions corresponding to the rank and the output size. Matrix $A$ is initialized with a mean of 0 and a standard deviation of 0.02, whereas matrix $B$ is initialized with zeros. These matrices are then multiplied together and added to the original weight matrix.
We then fine-tune only these LoRA parameters, while keeping the rest of the 
model unchanged.
In prefix tuning (F.3 of Figure~\ref{fig: method overview}), the initial domain information embedding serves as a starting point for further refinement. During training, both the the NN and the domain information embedding of the source domain are trained. For target adaptation, 
we switch the domain information embedding from the source to the
target domain. 
The NN is kept frozen, and only the domain information embedding of the target domain is updated using these samples from the target domain. 
We refer to this process as prefix tuning. 

As shown in Table~\ref{table:overview}, we use different hyperparameters for 
target adaptation. Detailed hyperparameters are provided in 
Appendix~\ref{sec: appendix hps}, and the 
hyperparameter selection process is discussed in Section~\ref{sec: testbed setup}.

\section{Numerical Experiments}
\label{section: experiments}

\begin{table}[t]
\caption{Details of datasets used in this work. ``\# Source$\rightarrow$Target Pair'' denotes the number of distribution shift pair for each dataset, and we consider the natural \emph{spatial} shift between US states.}
\label{table:overview}
\resizebox{\textwidth}{!}{
\begin{tabular}{clcccccc}
\toprule
\#ID & Dataset                                                                                        & \#Samples & \#Features & Outcome                   & \#Source Domains & \#Target Domains & \#Source$\rightarrow$Target Pair                                                                 \\ \midrule
1 & \texttt{ACS Income}        & 1.60M & 9          & Income$\geq$50k           & 51 (US States)        & 50 (US States) & 2550                                                                    \\
2 & \texttt{ACS Mobility}      & 621K   & 21         & Residential Address       & 51 (US States)        & 50 (US States)& 2550                                                                       \\
3 & \texttt{ACS Pub.Cov}       & 1.12M & 18         & Public Ins. Coverage      & 51 (US States)        & 50 (US States)& 2550             \\\bottomrule                                                      
\end{tabular}}
\end{table}

\begin{figure}[t]
\centering\includegraphics[width=0.9\textwidth]{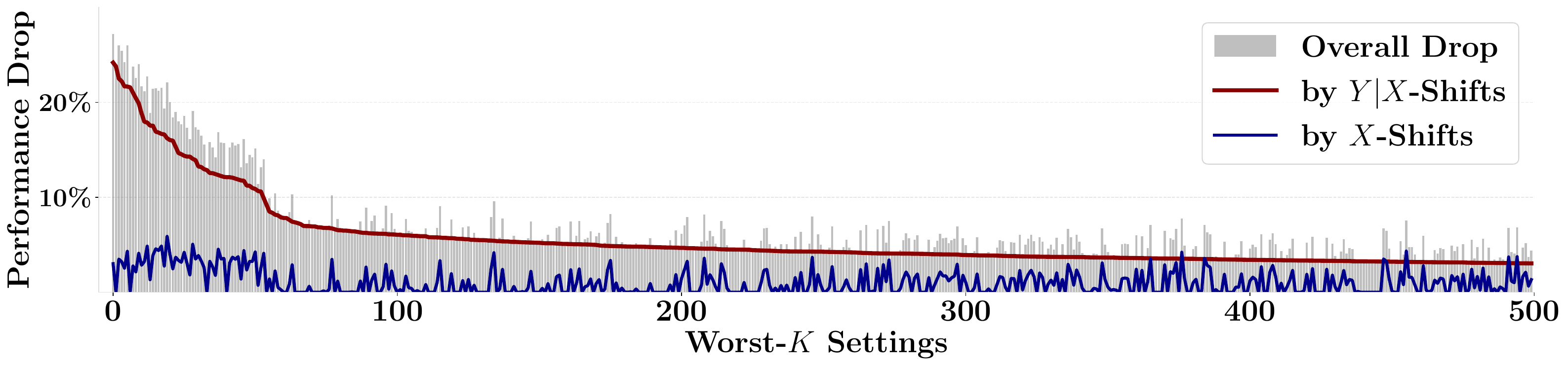}
\caption{Shift pattern analysis. For the 2550 source$\rightarrow$target distribution shift pairs in \texttt{ACS Income} dataset, we attribute the performance drop for each source$\rightarrow$target pair into $Y|X$-shifts (red curve) and $X$-shifts (blue curve), and sort all pairs according to the drop introduced by $Y|X$-shifts. We draw the \emph{worst-500} settings in each dataset, and the decomposition method used here is DISDE~\citep{CaiNaYa23} with XGBoost as the reference model. Results on other datasets are in~\Cref{fig-appendix:overall_decomposition}.}
\label{fig:overall_decomposition}
\end{figure}

In this section, we conduct a thorough investigation of \textbf{7650} natural shift settings (source $\rightarrow$ target domain) in 3 tabular datasets over \textbf{261,000} model configurations and summarize the observations.
Our findings highlight the potential of incorporating LLM embeddings to enhance the generalization ability in tabular data prediction tasks.

\subsection{Testbed Setup}\label{sec: testbed setup}

\paragraph{Dataset} In this work, we use the ACS dataset~\citep{ding2021retiring} derived from the US-wide ACS PUMS data, where the goal is to predict various socioeconomic factors for individuals.
\begin{itemize}
\item \texttt{ACS Income}: The goal is to predict whether an individual’s income is above \$50K based on individual features. We filter the dataset to only include individuals above 16 years old with usual working hours of at least 1 hour per week in the past year, and an income of at least \$100. 
\item \texttt{ACS Mobility}: The goal is to predict whether an individual has the same residential address as one year ago.
We filter the dataset to only include individuals between the ages of 18 and 35, which increases the difficulty of the prediction task.
\item \texttt{ACS Public Coverage} (abbr. as \texttt{ACS Pub.Cov}):
The goal is to predict whether an individual has public health insurance. We focus on low-income individuals who are not eligible for medicare by filtering the dataset to only include individuals under the age of 65 and with an income of less than \$30,000. 
\end{itemize}
The details of datasets are summarized in \Cref{table:overview}.

\paragraph{Shift Pattern Analysis} Before benchmarking, we first analyze the shift patterns among the 2550 source$\rightarrow$target pairs in each dataset.
Specifically, we utilize DISDE~\citep{CaiNaYa23} (reference model as XGBoost) to decompose the performance degradation from the source domain to the target into two parts: (a) $Y|X$ (concept)-shifts and (b) $X$ (covariate)-shifts.
By utilizing tailored shift patterns, we can conduct an in-depth analysis of where the strength of LLM embeddings lies.
As shown in~\Cref{fig:overall_decomposition}, we sort all pairs according to the strength of $Y|X$-shifts, where we find that the natural spatial shifts are mainly comprised of $Y|X$-shifts.
These findings broaden the scope of the analysis in \texttt{WhyShift}~\citep{LiuWaCuNa23} by examining 7,650 shift pairs, a significant increase from the 169 pairs studied in the original work.

\paragraph{Algorithms} As introduced in~\Cref{sec:methods}, we compare various methods that incorporate LLM embeddings into tabular data prediction, including  different finetuning methods (no finetuning, finetuing on full parameters, and low rank adaptation (LoRA)) and different embeddings (w/ or w/o extra information).
Besides, in order to fully compare the performances, we also include a wide range of learning strategies that perform on \textsf{Tabular} features, including basic models (LR, SVM, NN), tree ensembles (XGB, LGBM, GBM), robust methods (KL-DRO, $\chi^2$-DRO, Wasserstein DRO, CVaR-DRO, and Unified-DRO), and in-context learning methods (TabPFN, GPT-4o-mini).   
All methods are summarized in~\Cref{tab:overview-methods}.

\paragraph{Experiment Setup} 
We conduct experiments with more than \textbf{261,000} model configurations on 2550 source$\rightarrow$target shift pairs in \texttt{ACS Income}, \texttt{ACS Mobility}, and \texttt{ACS Pub.Cov} datasets respectively (\textbf{7650} settings in total).
For each source$\rightarrow$target shift pairs, we randomly sample $20,000$ labeled data from the source and target domain respectively, as the training and test dataset. 
We evaluate the model trained on the source domain, with or without target adaptation, and report the \emph{Macro F1 score} on the testing dataset. 

\begin{figure}[t]
\centering
\includegraphics[width=\textwidth]{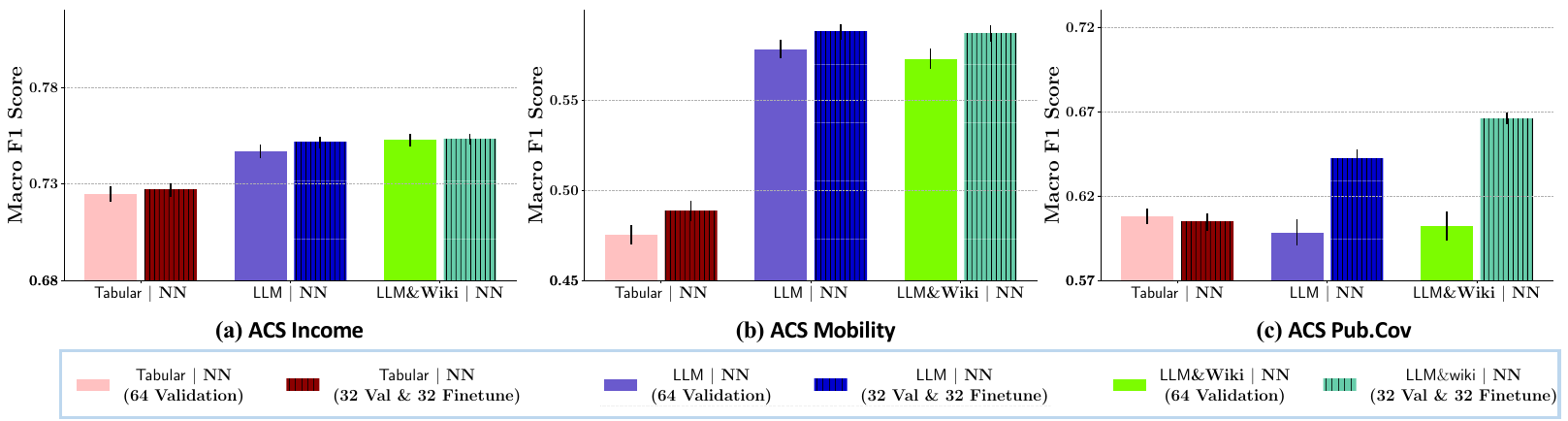}
\caption{Comparison between different allocation of target samples into validation and finetuning. We report the average Macro F1 Score over the worst-500 settings.}
\label{fig:allocation}
\end{figure}

Given the numerous training hyperparameters---learning rate, number of training epochs, hidden layer dimension, dropout ratio---we use a validation set of 32 randomly sampled labeled target domain samples to choose the optimal training hyperparameters, based on the highest F1 score in the validation dataset.
Since our metric is Macro F1 score, the validation set is set as balanced between positive and negative classes.
Note that the hyperparameter selection is \emph{near}-oracle, as it leverages target samples, albeit in a limited quantity.
When doing finetuning, we sample another 32 labeled target samples to finetune the model. 
And we use the same 32-sample validation dataset (for training hyperparameter selection) to select the target adaptation hyperparameters that yield the best Macro F1 score.
Note that our testbed allows flexible sample sizes for training, validation, testing, and finetuning.
Additionally, we perform an ablation study on different allocations of the overall target samples in validation and finetuning (see~\Cref{fig:allocation}).
See details on hyperparameters in the~\Cref{sec: appendix hps}.

\subsection{Primary Findings}\label{sec: primary finding}
We begin by presenting the key observations from our results. 
In addition to the metric~\eqref{equ:optimal-ratio} introduced in~\Cref{section: introduction}, we report performance metrics averaged over the source-target pairs.

\begin{figure}[t!]
\centering
\subfloat[\texttt{ACS Income}]{\includegraphics[width=0.32\textwidth]{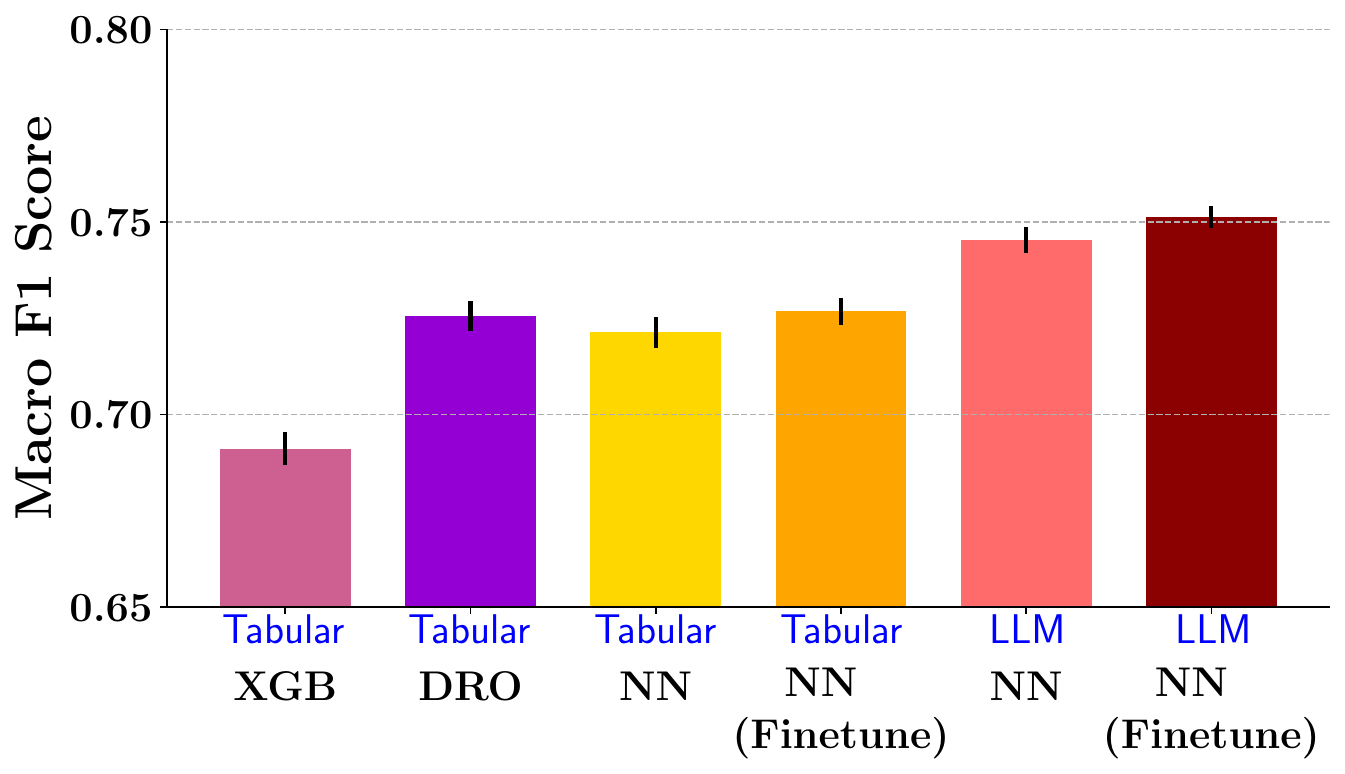}
\label{fig1:overall_subfig1}}
\subfloat[\texttt{ACS Mobility}]{\includegraphics[width=0.32\textwidth]{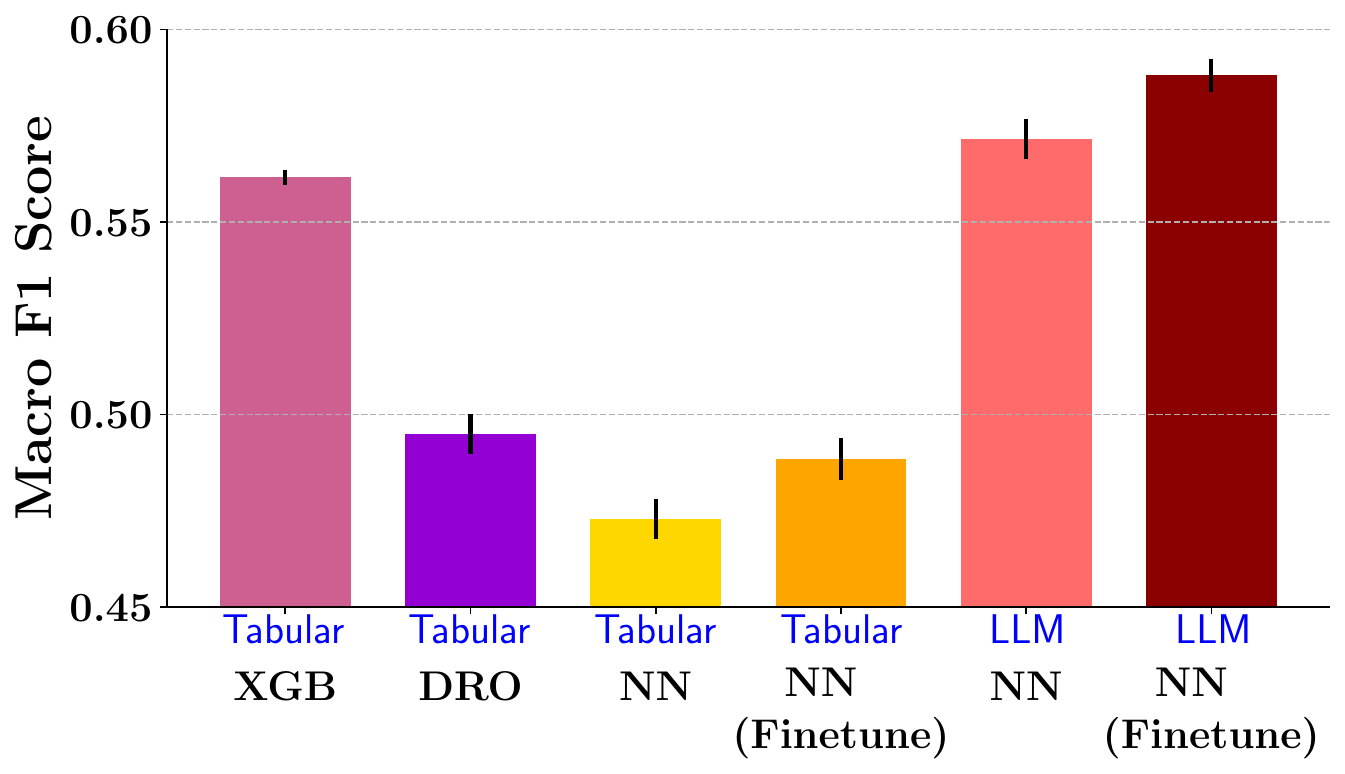}\label{fig1:overall_subfig2}}
\subfloat[\texttt{ACS Pub.Cov}]{\includegraphics[width=0.32\textwidth]{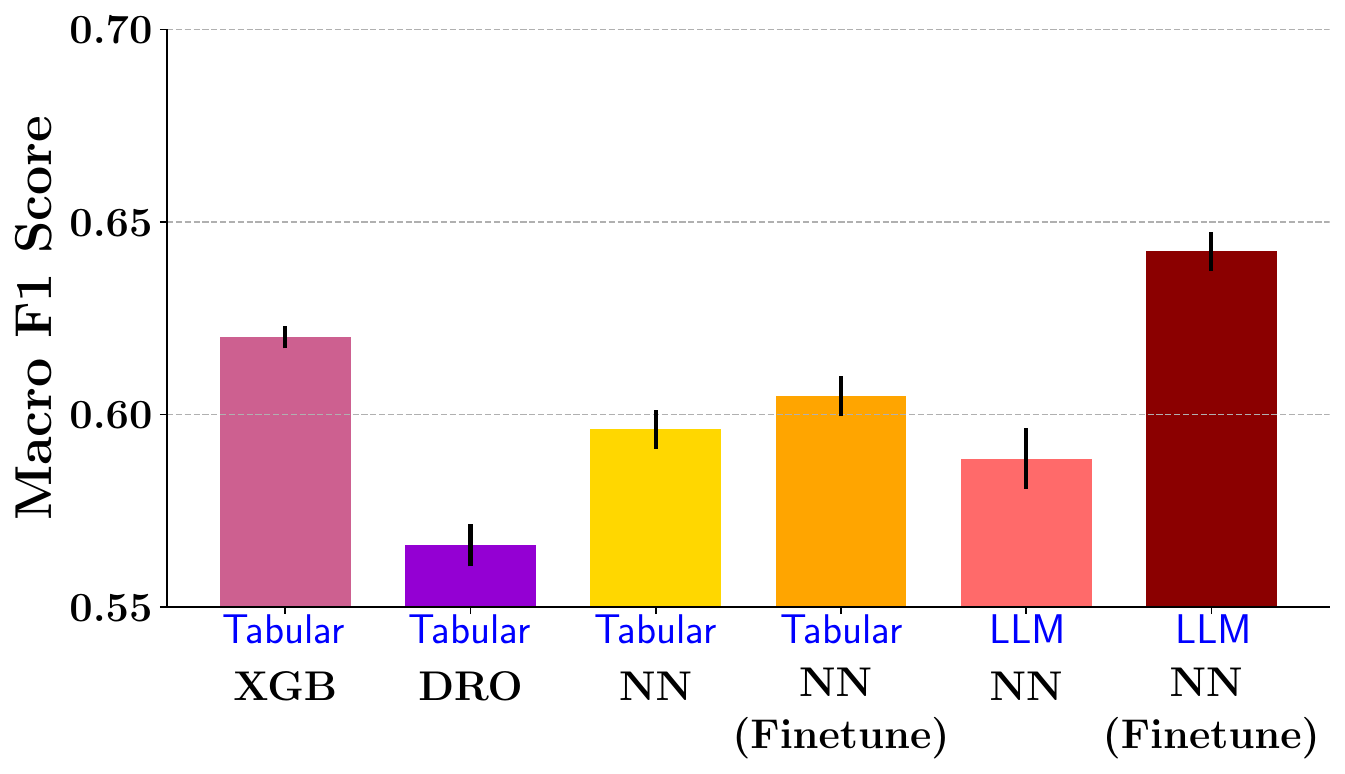}\label{fig1:overall_subfig4}}
\caption{Average Macro F1 Score over the worst-500 settings. For each dataset, we sort the 2550 settings according to the magnitude of $Y|X$-shifts and select the \textbf{worst-500} settings. We calculate the average Macro F1 Score for each method. For all methods, we select the best hyper-parameters of the basic model according to 32 samples from the target domain. We use CVaR-DRO based on NN here to represent DRO methods. For finetuning methods, we use an additional 32 target samples for finetuning; recall Section~\Cref{sec: testbed setup}.}
\label{fig:overall}  
\end{figure}

\paragraph{LLM embeddings improve performance, but when applied alone do not consistently outperform tree-ensembles.}
To better assess the generalization ability when incorporating LLM embeddings, we select the worst 500 settings (out of 2,550 total settings per dataset) based on the severity of $Y|X$-shifts and report the average Macro F1 Score in \Cref{fig:overall}. Each bar represents the average result across these worst 500 settings, characterized by the most severe $Y|X$-shifts. 
Thus, even a 1pp improvement is significant, as it implies consistent gains of about 1pp across each of the worst 500 settings.

Comparing ``NN on LLM embeddings" to ``NN on tabular features" (with the backbone model fixed as NN), we observe LLM embeddings significantly enhance generalization under distribution shifts on the \texttt{ACS Income} and \texttt{ACS Mobility} datasets, with average improvements of 2.4pp and 9.9pp. Notably, ``NN on LLM embeddings" even outperforms XGBoost under strong distribution shifts on these datasets. This demonstrates the potential of LLM embeddings in tabular data prediction, where they can contribute to more generalizable models.

A different trend is observed on the \texttt{ACS Pub.Cov} dataset, where the inclusion of LLM embeddings results in a performance drop for NN models. This suggests that simply incorporating LLM embeddings does not always resolve distribution shift issues; their effectiveness may vary across datasets, particularly depending on whether the LLM embeddings provide additional relevant information for the specific prediction task.

\begin{figure}[t!]
\centering
\subfloat[\texttt{ACS Income}]{\includegraphics[width=0.33\textwidth]{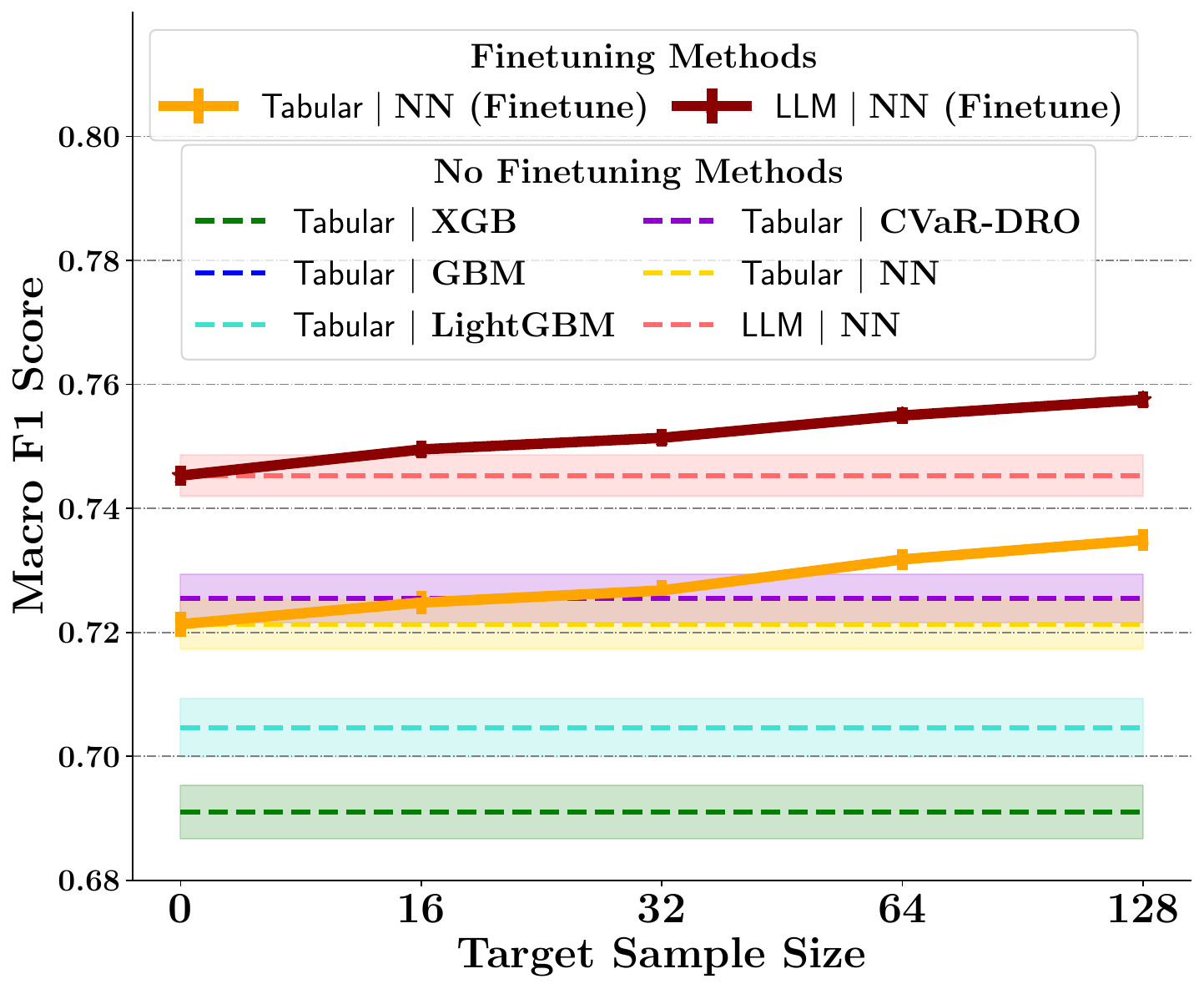}
\label{fig2:overall_subfig1}}
\subfloat[\texttt{ACS Mobility}]{\includegraphics[width=0.33\textwidth]{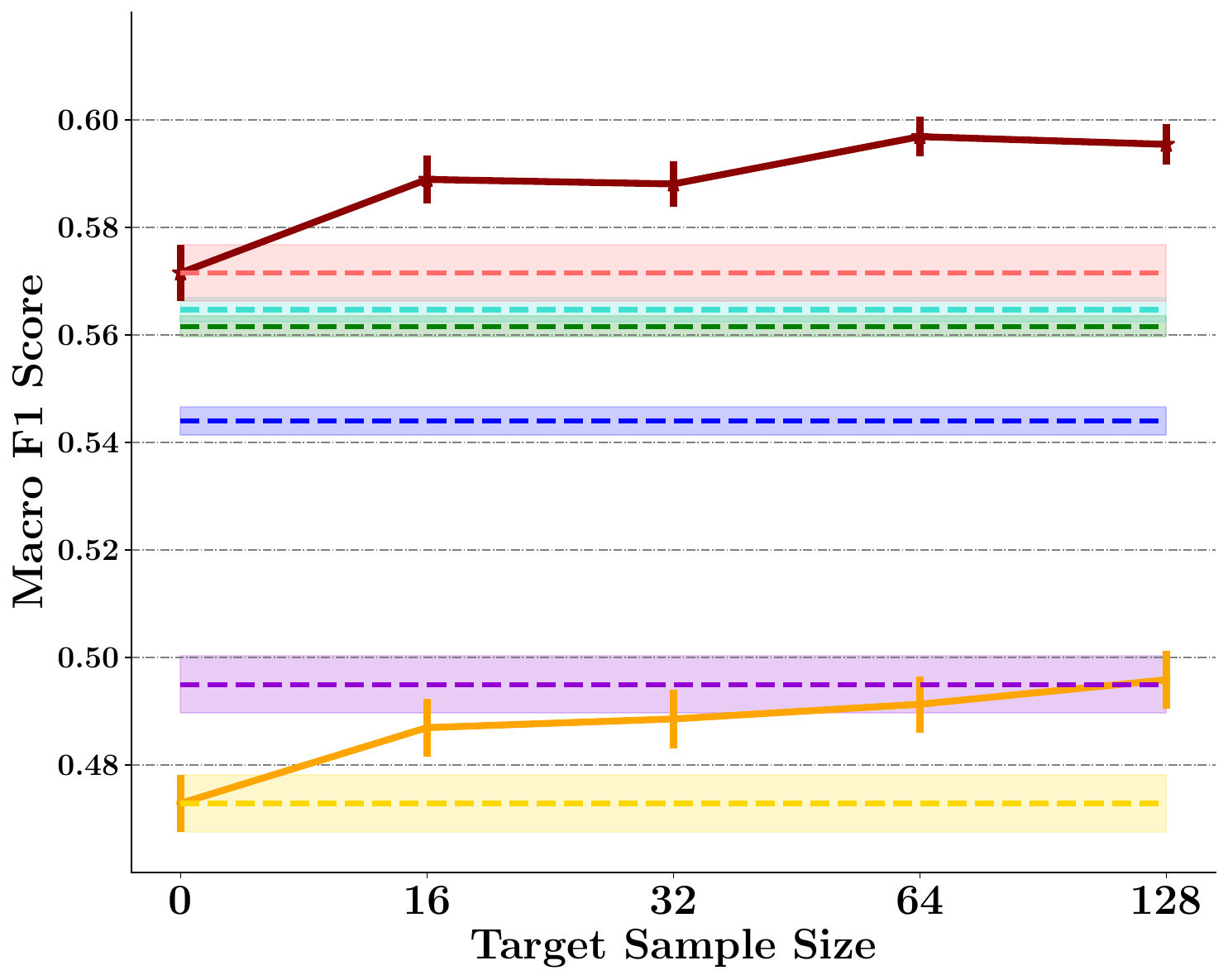}\label{fig2:overall_subfig2}}
\subfloat[\texttt{ACS Pub.Cov}]{\includegraphics[width=0.33\textwidth]{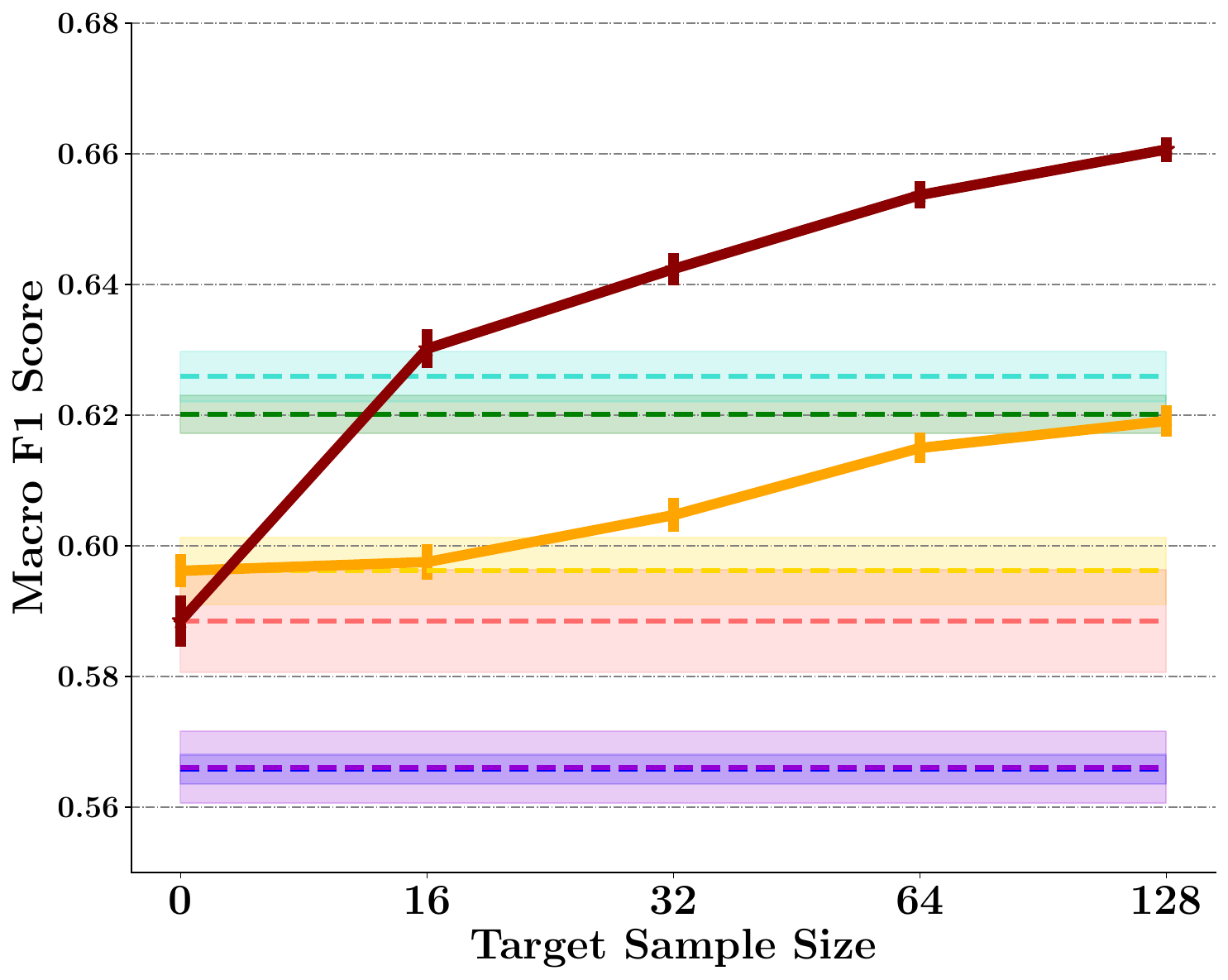}\label{fig2:overall_subfig4}}
\caption{Average Macro F1 Score over Worst-500 settings with different \#target samples used for finetuning. Dotted lines represent methods that do not require finetuning, while solid lines represent finetuning methods. All three figures share the same legend, and all results use 32 labeled target samples as validation dataset.}
\label{fig:target-number}
\end{figure}

\begin{figure}[t!]
\centering
\subfloat[\texttt{ACS Income} (no extra)]{\includegraphics[width=0.32\textwidth]{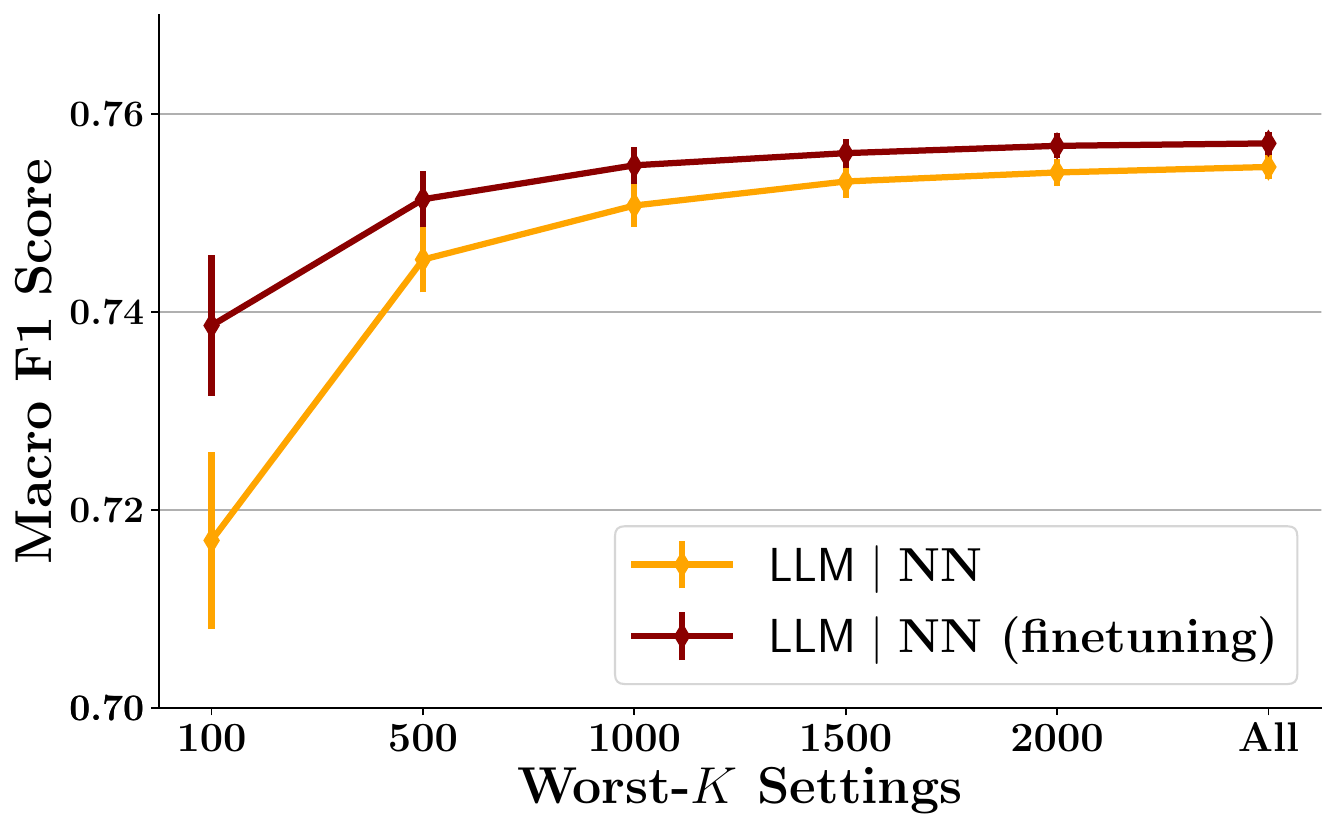}}
\subfloat[\texttt{ACS Mobility} (no extra)]{\includegraphics[width=0.32\textwidth]{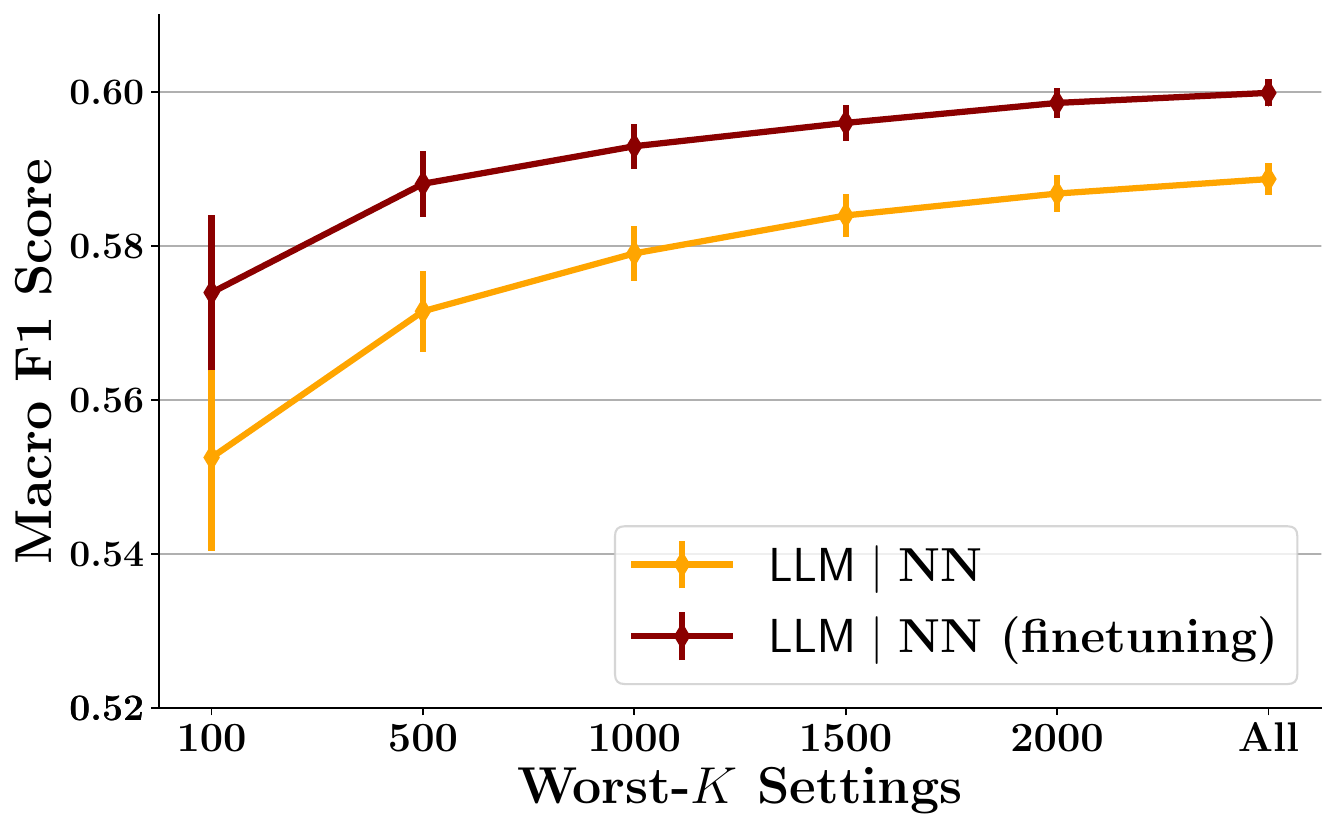}}
\subfloat[\texttt{ACS Pub.Cov} (no extra)]{\includegraphics[width=0.32\textwidth]{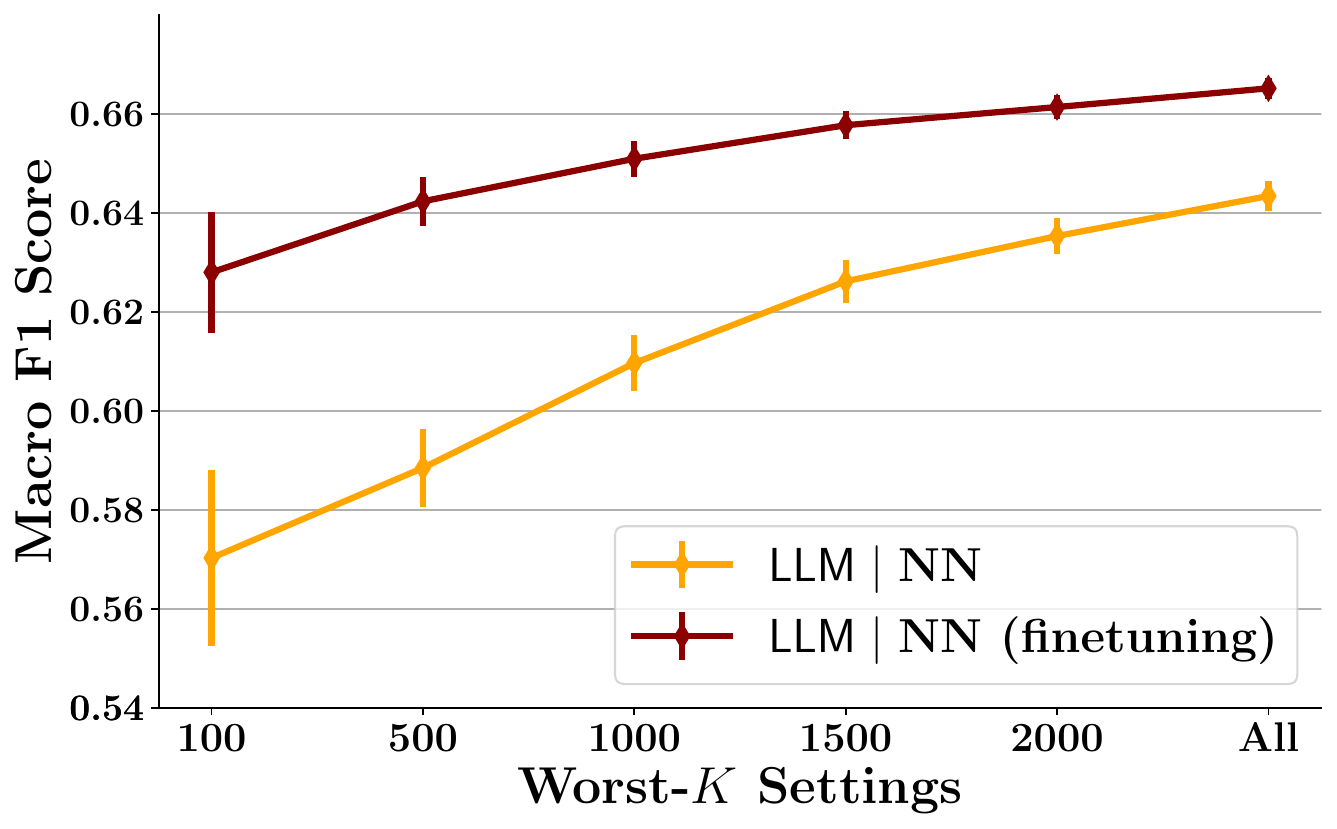}}\\
\subfloat[\texttt{ACS Income} (Wikipedia)]{\includegraphics[width=0.32\textwidth]{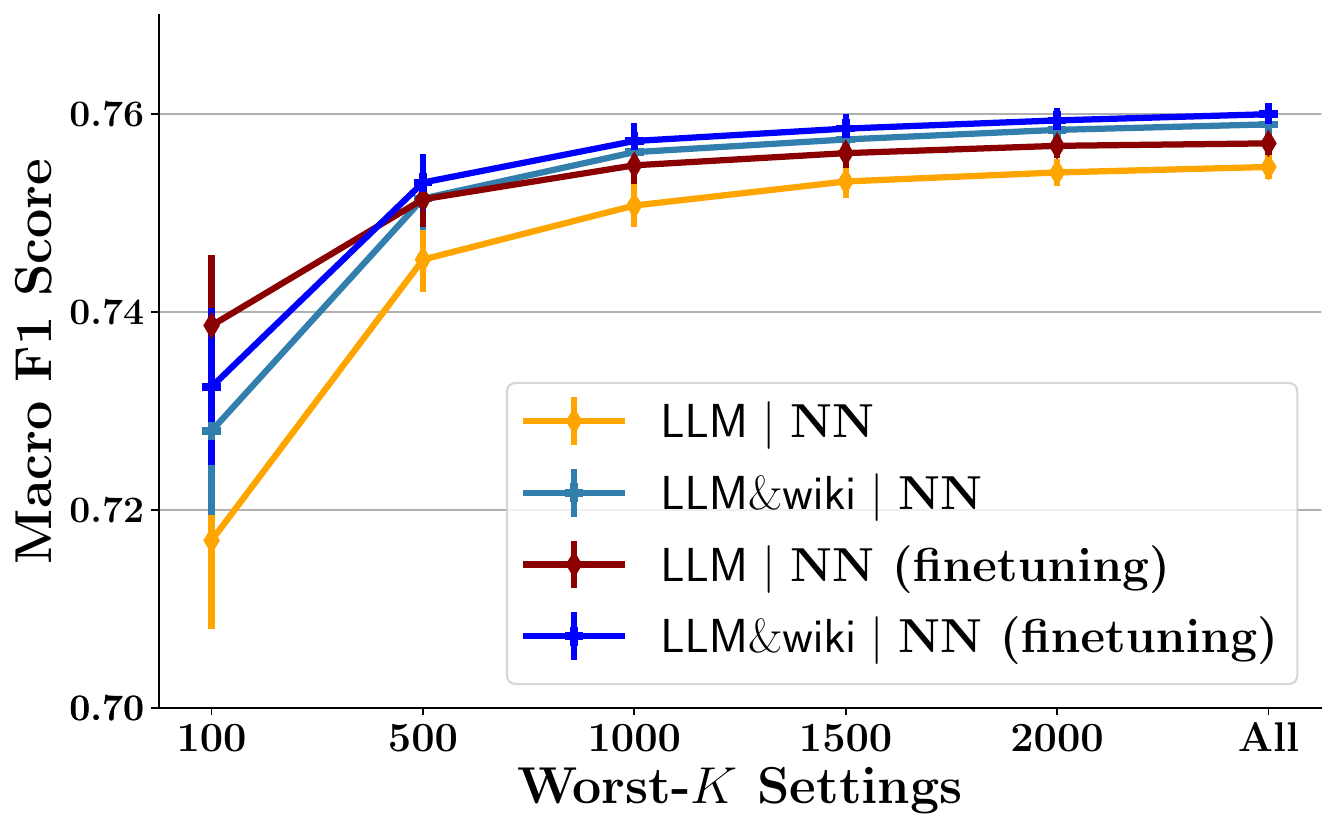}}
\subfloat[\texttt{ACS Mobility} (Wikipedia)]{\includegraphics[width=0.32\textwidth]{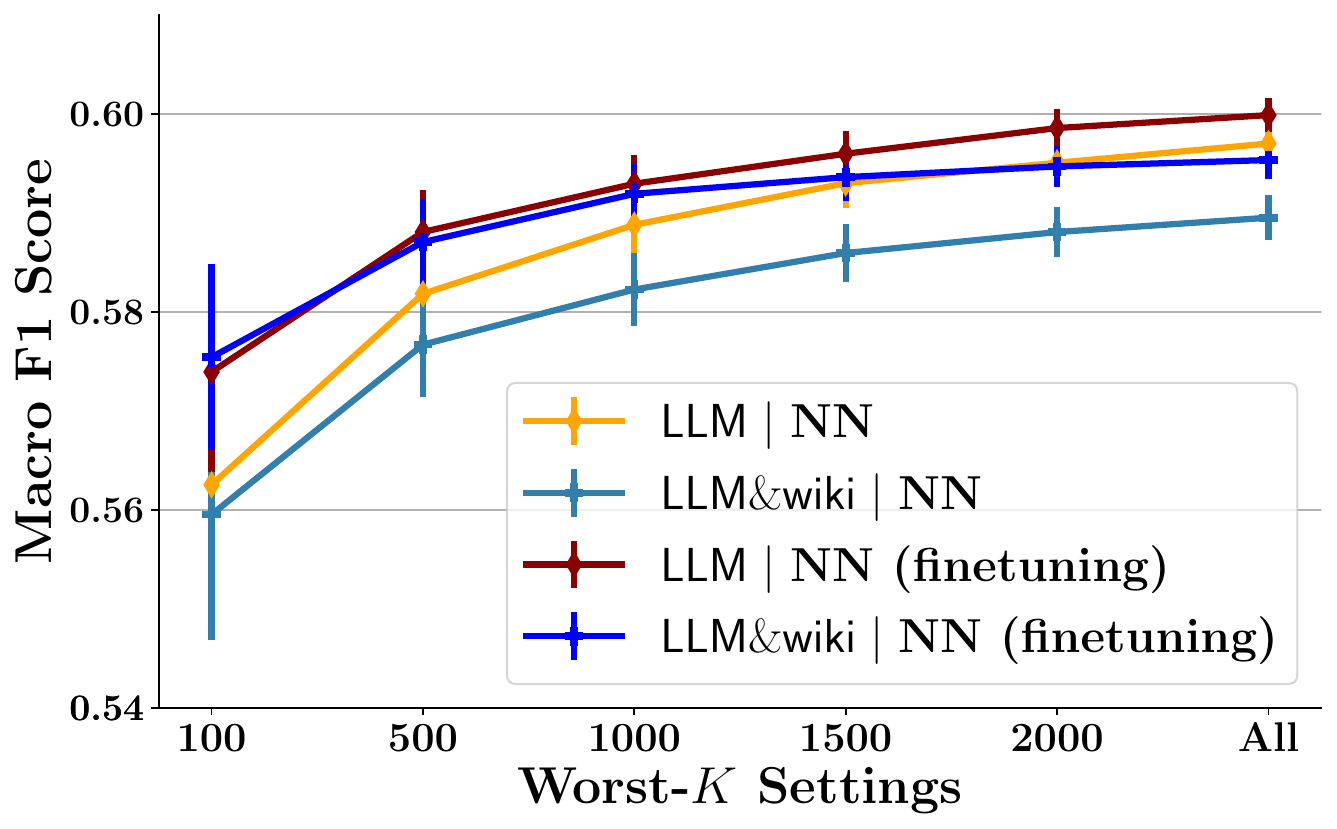}}
\subfloat[\texttt{ACS Pub.Cov} (Wikipedia)]{\includegraphics[width=0.32\textwidth]{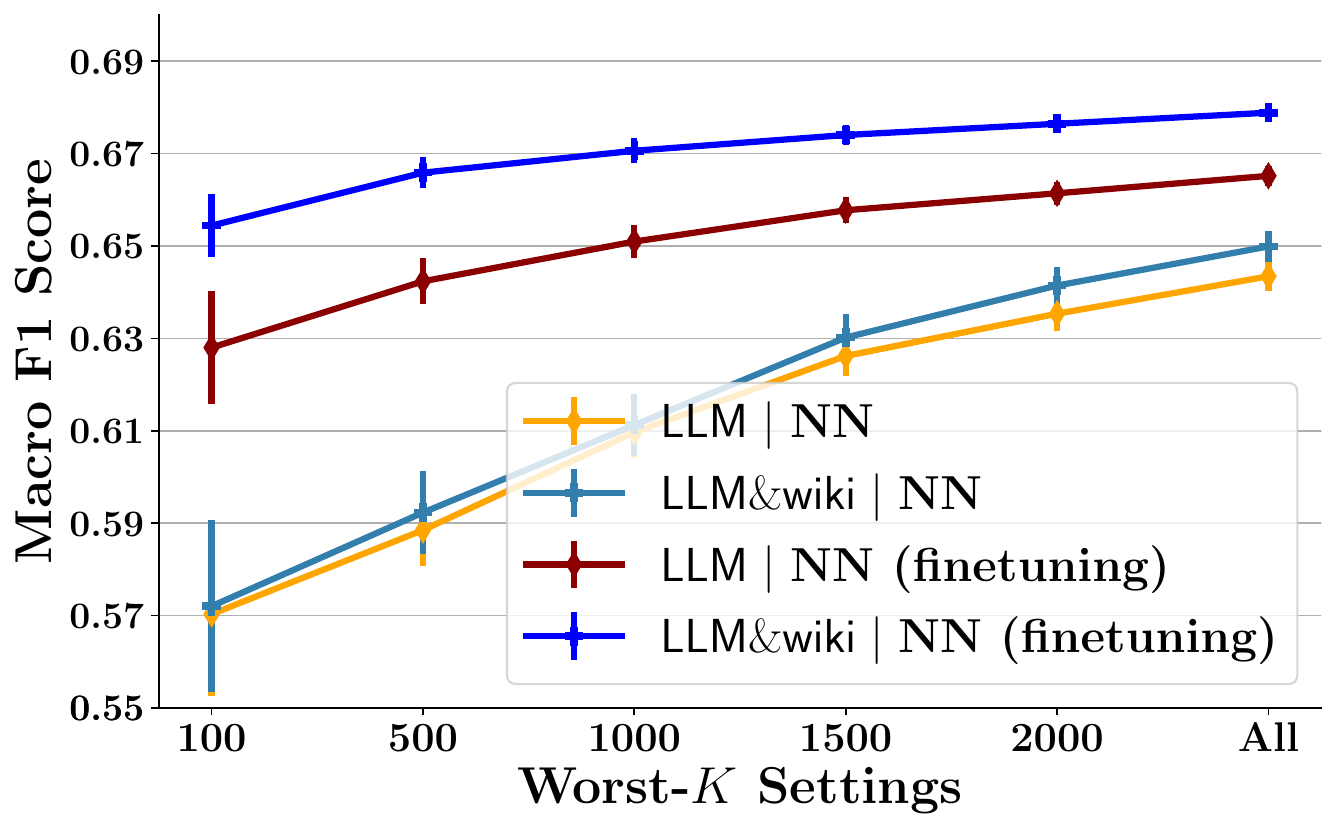}}\\
\subfloat[\texttt{ACS Income} (GPT-4)]{\includegraphics[width=0.32\textwidth]{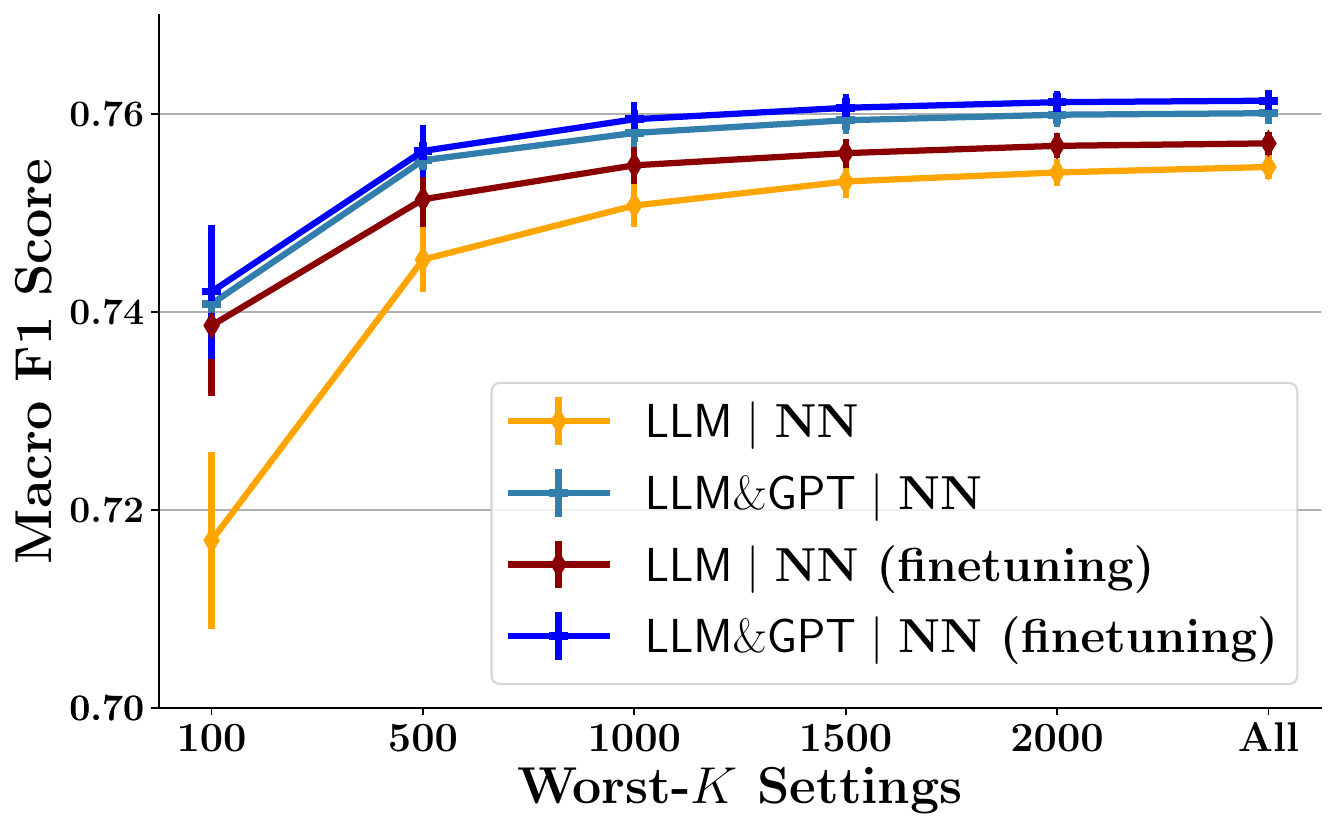}}
\subfloat[\texttt{ACS Mobility} (GPT-4)]{\includegraphics[width=0.32\textwidth]{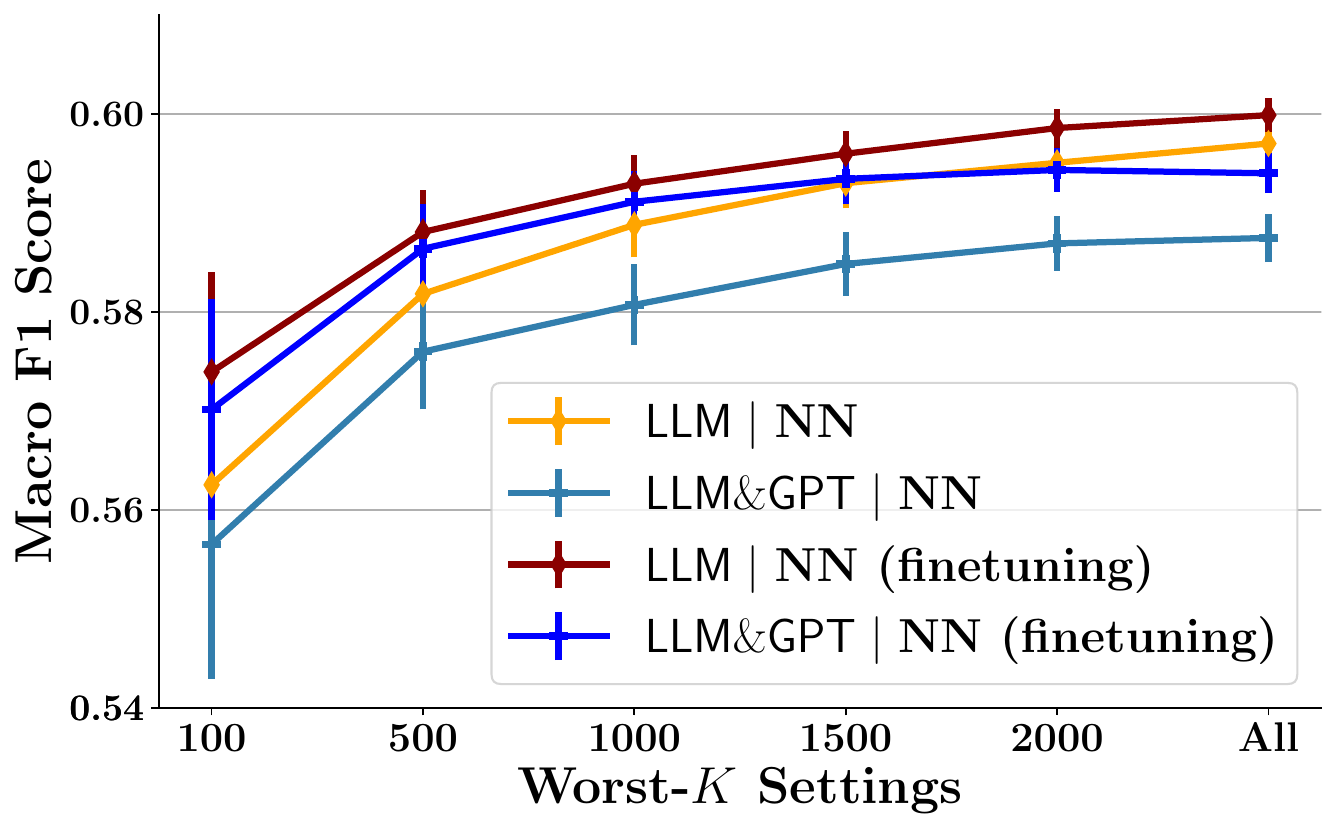}}
\subfloat[\texttt{ACS Pub.Cov} (GPT-4)]{\includegraphics[width=0.32\textwidth]{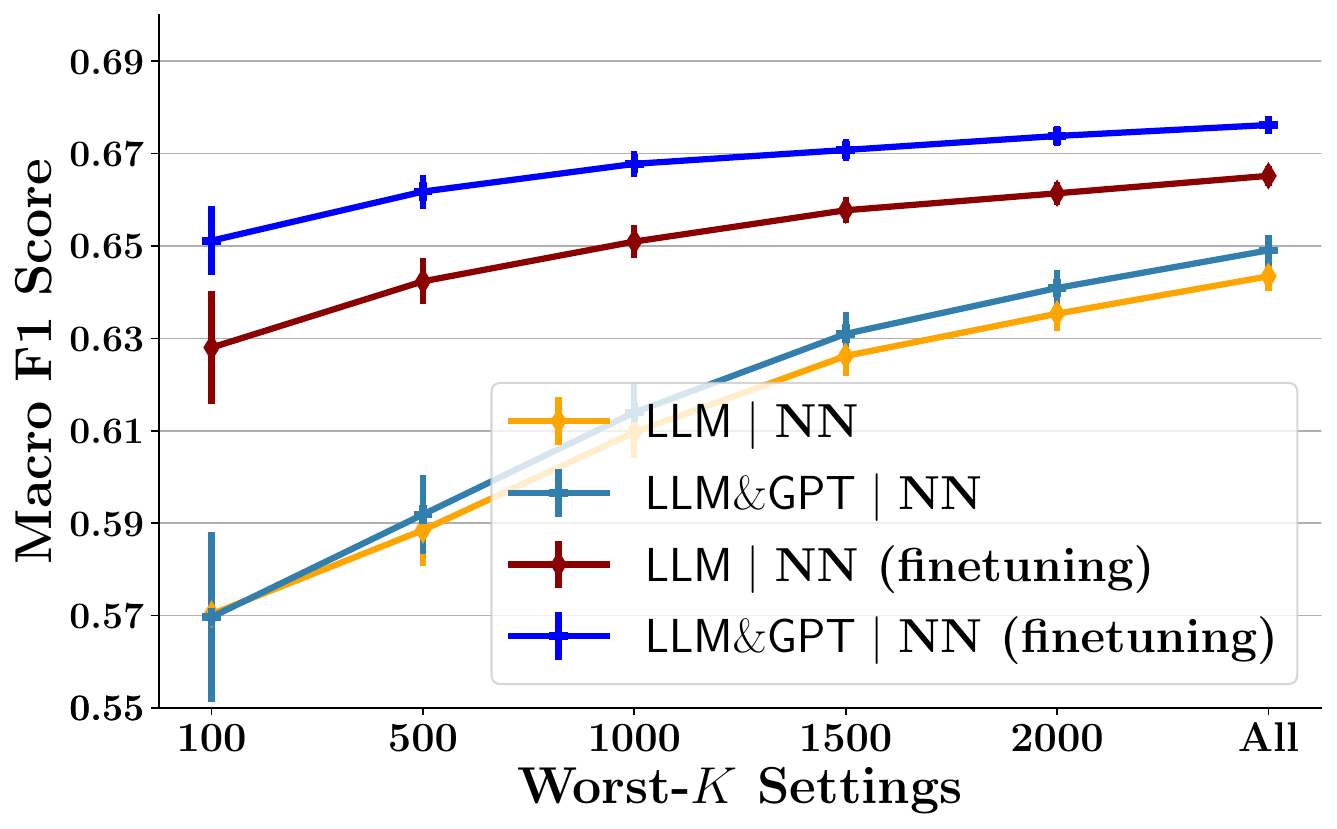}}\\
\subfloat[\texttt{ACS Income} (In-context)]{\includegraphics[width=0.32\textwidth]{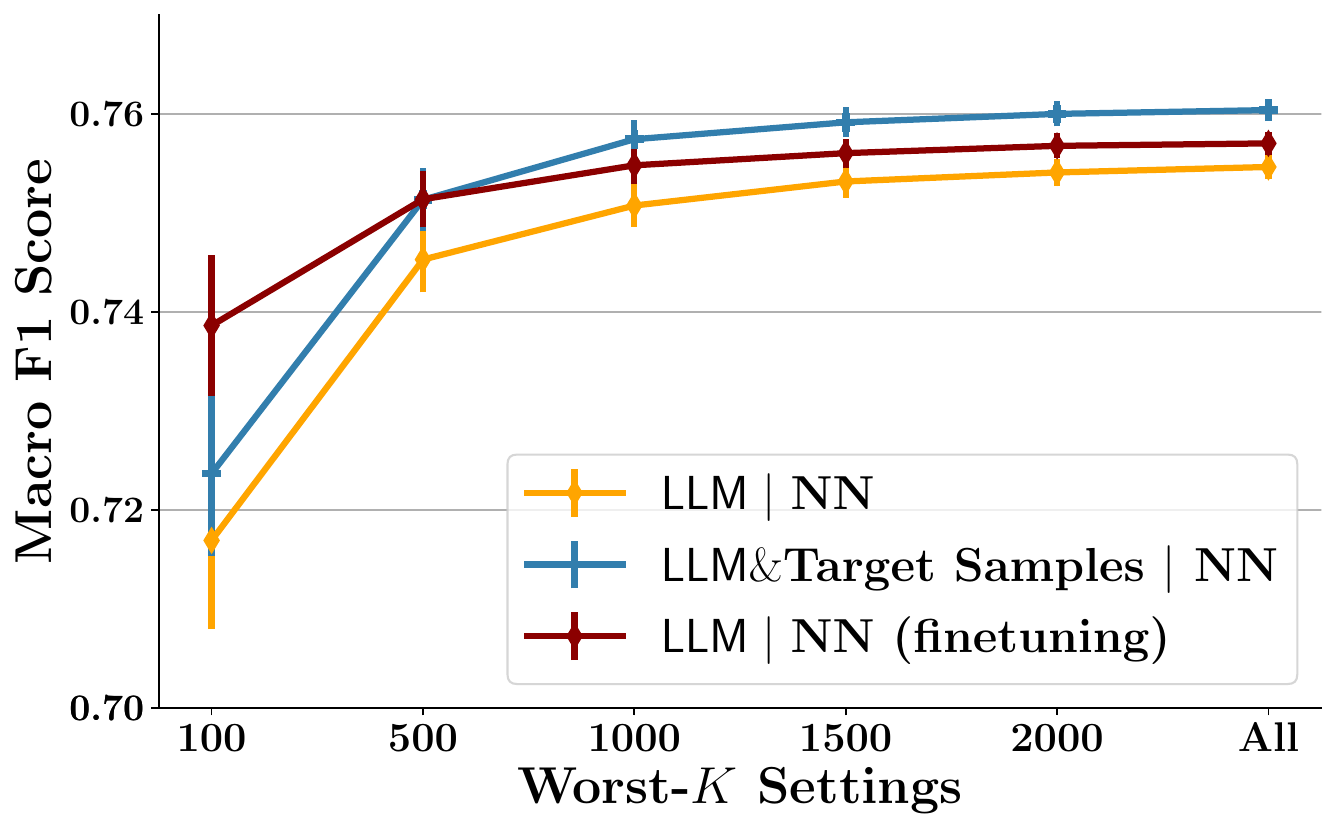}}
\subfloat[\texttt{ACS Mobility} (In-context)]{\includegraphics[width=0.32\textwidth]{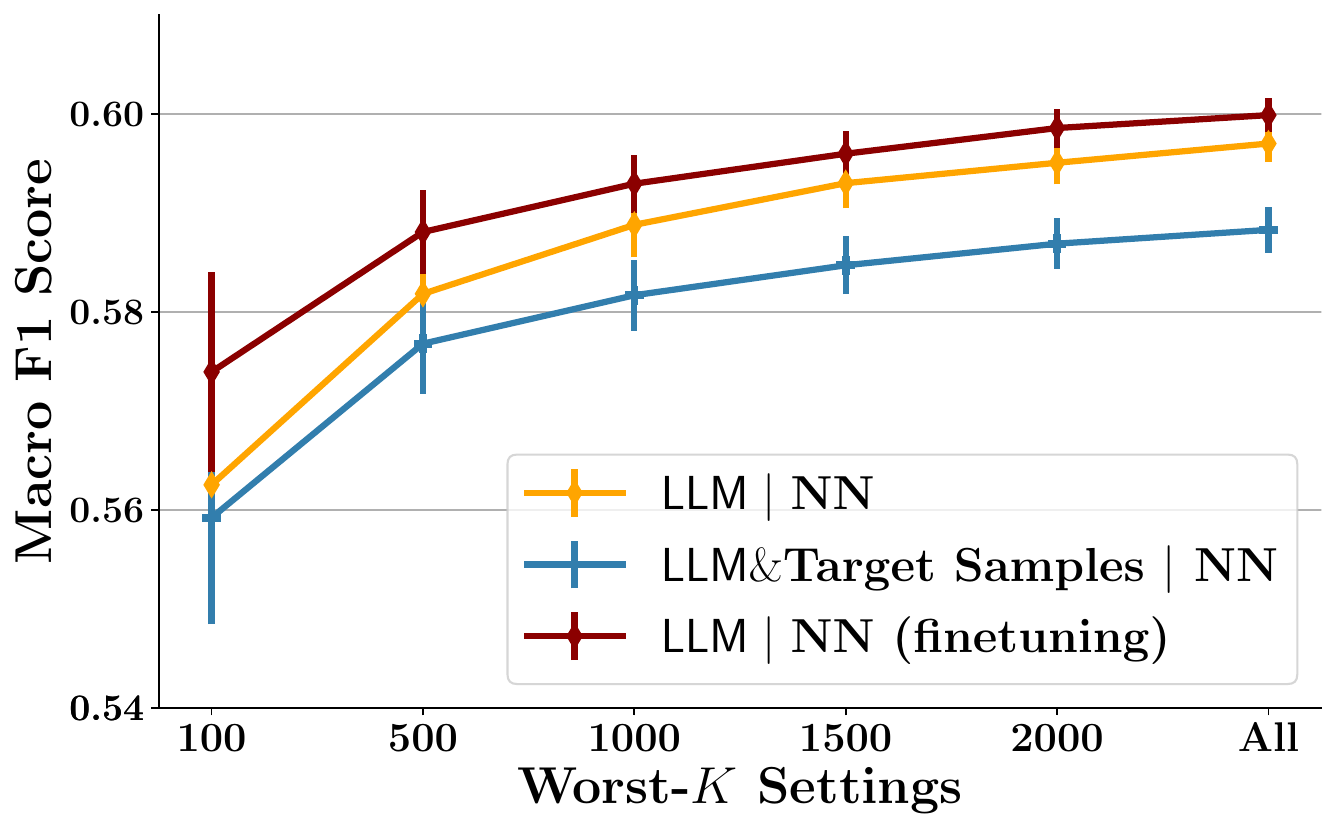}}
\subfloat[\texttt{ACS Pub.Cov} (In-context)]{\includegraphics[width=0.32\textwidth]{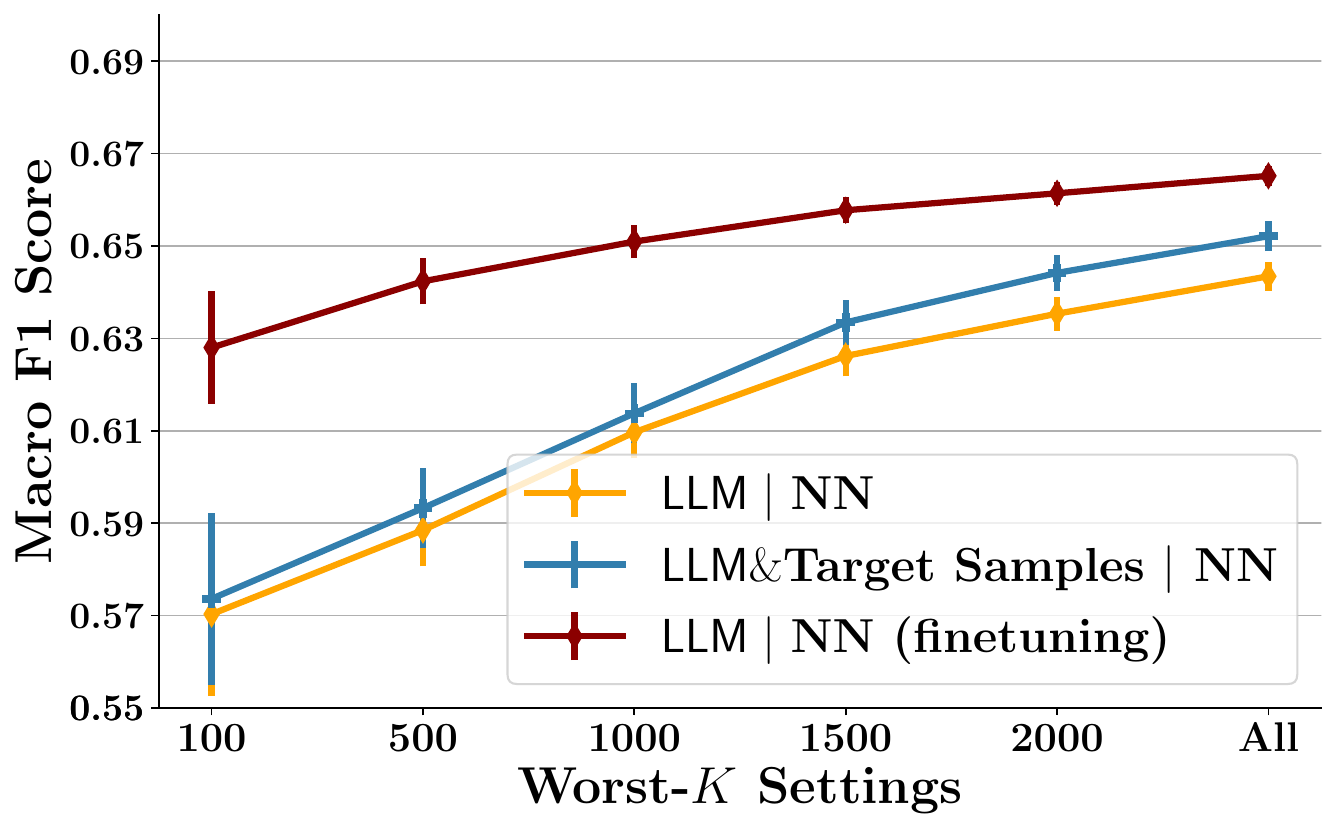}}\\
\caption{Average performance over the worst-$K$ pairs. For each dataset, we sort the 2550 pairs according to the magnitude of $Y|X$-shifts and select the worst-$K$ settings ($K\in \{100, 500,1000,1500,2000,2550\}$). For methods requiring finetuning, we use 32 target samples here. (a)-(c): NN based on LLM embeddings without extra information; (d)-(f): With domain information from Wikipedia; (g)-(i): With domain information generated by GPT-4; (j)-(l): With domain information from target-sample serialization (F.1 of \Cref{fig: method overview}).}
\label{fig:curve}  
\end{figure}

\paragraph{A small number of target samples can make a big difference.}
While incorporating LLM embeddings doesn't always yield improvements, we find that even a small number of target samples can have a significant impact.
As shown in \Cref{fig:overall}, finetuning the ``NN on LLM embeddings" model with just 32 target samples significantly improves the average performance across the worst 500 settings for both \texttt{ACS Mobility} and \texttt{ACS Pub.Cov}. Similar trends are observed for other worst-$K$ settings, as shown in~\Cref{fig:curve} (a)(b)(c) to come. 
Notably, for the \texttt{ACS Pub.Cov} dataset, where LLM embeddings alone provided no improvement, finetuning with only 32 target samples leads to a 5.4pp gain, even surpassing XGBoost by 2.2pp.
This highlights the adaptability of LLM embeddings, making them a promising tool for harnessing their power across various downstream real-world tasks.

Furthermore, in~\Cref{fig:target-number}, we illustrate how the performance of finetuning methods varies with different numbers of target samples used for finetuning. 
Our conclusions hold true regardless of the number of target samples used for finetuning.

\begin{figure}[t]
\centering
\includegraphics[width=\textwidth]{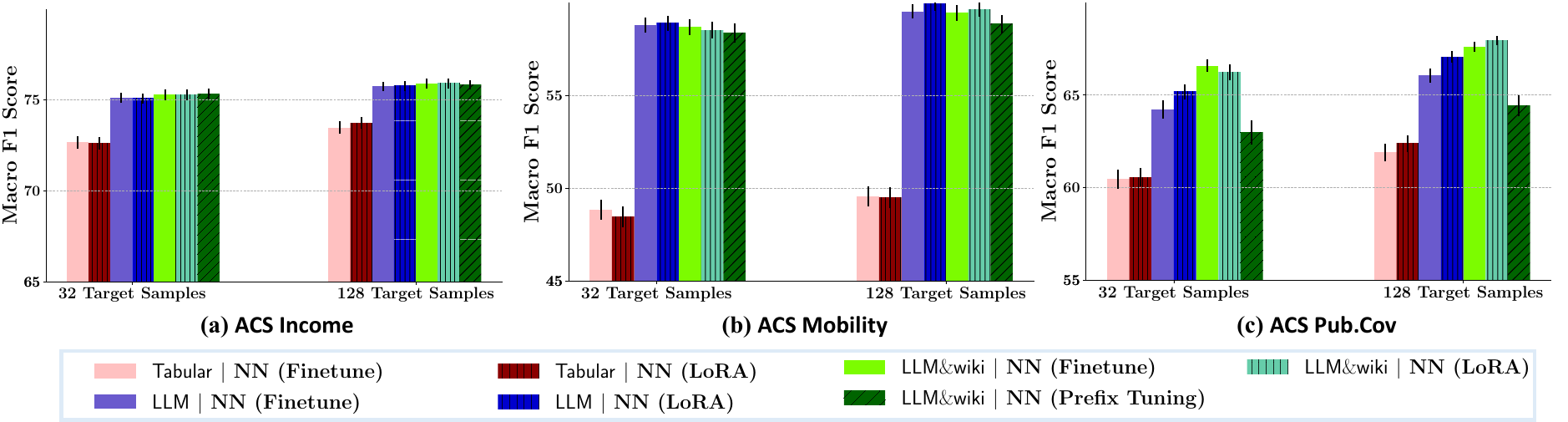}
\caption{Comparison between full-parameter finetuning, LoRA, and prefix tuning. We show the average Macro F1 Score over the worst-500 settings.}
\label{fig:lora}
\end{figure}

\paragraph{The performance gain brought by finetuning with target samples is larger under stronger $Y|X$-shifts.}
As shown in~\Cref{fig:curve} (a)-(c) (LLM $|$ NN v.s. LLM $|$ NN (finetuning)), for NN using LLM embeddings, finetuning with 32 target samples yields an average performance gain of 5.8pp across the worst-100 settings on \texttt{ACS Pub.Cov}, compared to no finetuning. This is notably higher than the 1.2pp average gain observed across all 2550 settings (\textbf{4.8 times}). 
Similarly, for \texttt{ACS Income}, it is about \textbf{9 times}.

\subsection{Auxiliary Findings}\label{sec: aux finding}
In addition to the primary findings, we have several other noteworthy observations.

\paragraph{``Right'' domain information matters.} 
In~\Cref{fig:curve} (d)-(l), we study how additional domain information --- from Wikipedia, GPT-4, or 32 target samples (via in-context domain information, F.1 of \Cref{fig: method overview})
impacts F1 scores in both non-adaptation and target adaptation scenario. We do NOT observe a consistent trend across all three datasets: additional domain information can either improve or reduce F1 scores. In the non-adaptation case, both LLM\&wiki $|$ NN and LLM\&GPT $|$ NN outperform LLM $|$ NN in \texttt{ACS Income} (d)(g), but underperform in \texttt{ACS  Mobility} (e)(h). When finetuning NNs, LLM\&wiki $|$ NN (finetuning) and LLM \& GPT $|$ NN (finetuning) beat LLM $|$ NN (finetuning) in \texttt{ACS Pub.Cov} (f)(i), but underperform in \texttt{ACSMobility}. Moreover, LLM\& Target Samples $|$ NN shows comparable performance with  LLM $|$ NN (finetuning) in \texttt{ACSIncome} (j), but underperform LLM $|$ NN (finetuning) in \texttt{ACS Mobility} and \texttt{ACS Pub.Cov} (k)(l), and even underperform LLM $|$ NN that has no target adaptation in \texttt{ACS Mobility} (k). 
Our results indicates that ``right" domain information can indeed improve tabular data classification under $Y|X$ shifts, yet identifying the best domain information requires non-trivial engineering efforts, which we leave for future work. 

For simplicity, we only compare LLM embeddings (without extra domain information) to LLM\&Wiki in the sequel, as LLM\&Wiki and LLM\&GPT exhibit similar trends.

\paragraph{Specific finetuning approaches are less crucial than expected.}
Given the large number of model parameters and limited labeled target samples, one might expect parameter-efficient methods like Low-Rank Adaptation (LoRA) and Prefix Tuning (F.3 of Figure~\ref{fig: method overview}) to offer a clear advantage. However, as shown in~\ref{fig:lora}, 
all methods significantly outperform the non-finetuned version when using LLM embeddings, 
and the choice of finetuning method appears less significant in our setting.

An exception is prefix tuning on the \texttt{ACS Pub.Cov} task, where performance was several percentage points lower. While this requires further investigation, our key takeaway is that under $Y|X$ shifts, 1) finetuning models using LLM embeddings can greatly enhance classification performance; 2)  popular finetuning methods yield comparable results.

\paragraph{Allocation of labeled target samples matters.}
In Figure~\ref{fig:allocation}, we compare two allocation schemes of $64$ labeled target samples. For the solid bars, all 64 samples are utilized as the validation dataset for hyperparameter selection. For the shaded bars,  we allocate 32 samples for validation and the remaining 32 samples for finetuning. 
Based on this, we initially explore and understand the impact of sample allocation.
For \texttt{ACS Mobility} and \texttt{ACS Pub.Cov}, target adaptation using LLM embeddings (the shaded bar) significantly outperforms the validation approach (the solid bar). However, for \texttt{ACS Income}, the improvement from target adaptation is only marginal.
This shows that although target adaptation is effective, its improvement is highly dependent on the specific distribution and $Y|X$ shift level. This raises further questions about how to optimally allocate resources. Although our findings endorse considering target adaptation, identifying the best allocation strategy is left as future work.

\section{Discussion based on theoretical insights}
\label{sec: theory}

We take a brief examination of the theoretical insights that may lie behind our empirical findings. While standard generalization bounds are vacuous for neural networks, we nevertheless find that theoretical results from domain adaptation provide a useful starting point for understanding why finetuning LLM embeddings with small target samples can result in superior generalization performance under substantial $Y|X$ shifts, particularly in comparison to tabular features.

Let $\phi(X)$ denote a feature map and $Y$ a binary label. We let $P$ and $Q$ be the source and target distributions.  We consider a  model class  \( \mathcal{H} \) of VC dimension $d$. For any model $h\in \mathcal{H}$, we use $\epsilon_{P}(h) := \mathbb{E}_{P}[\mathbb{I}(h(\phi(X)) \neq Y)]$  to denote the expected 0-1 loss on the source domain and $\hat{\epsilon}^m_{P}(h)$ its empirical counterpart based on $m$ i.i.d. samples from $P$. We define $\epsilon_{Q}(h)$ and $\hat{\epsilon}^m_{Q}(h)$ on the target. 

Although in practice we finetune on the target, we use a mixture problem as a rough approximation. Suppose that we have $(1-\beta)m$ i.i.d. samples from source domain $P$ and $(\beta m)$ i.i.d. samples from target domain $Q$. Let $\hat{h}_{\alpha, \beta} $ be the minimizer of the $\alpha$-weighted empirical error
$$\hat{h}_{\alpha, \beta} := \argmin_{h\in \mathcal{H}} \left\{\alpha \hat{\epsilon}^{\beta m}_{Q}(h) + (1 - \alpha) \hat{\epsilon}^{(1-\beta)m}_{P}(h)\right\}.$$
The following classical result from~\citet[Theorem 3]{ben2010theory} bounds
the generalization error on the target.
\begin{proposition}
For any $\delta\in(0,1)$, 
with probability at least $1-\delta$,
\begin{equation}
\label{equ:bound}
\begin{aligned}
    \epsilon_{Q}(\hat{h}_{\alpha, \beta})-
    \inf_{h \in \mathcal{H}} \epsilon_{Q}(h)\leq 4&\sqrt{\frac{\alpha^2}{\beta}+\frac{(1-\alpha)^2}{1-\beta}}\sqrt{\frac{2d\log(2(m+1))+2\log\frac{8}{\delta} }{m}} \\
    &+ 2(1-\alpha)\underbrace{d_{\mathcal H\Delta \mathcal H}(P_{X}, Q_{X})}_{X\text{-shifts}}+2(1-\alpha)\underbrace{
    \inf_{h\in\mathcal H}
    \left\{\epsilon_{P}(h)+\epsilon_{Q}(h)\right\}}_{Y|X\text{-shifts}},
\end{aligned}
\end{equation}
where $d_{\mathcal H\Delta\mathcal H}(\cdot,\cdot)$ denotes the $\mathcal H\Delta\mathcal H$-distance between two (marginal) distributions.
\end{proposition}

Since we use limited (32) labeled target samples for finetuning, $\beta$ is small. 
Also, we use a smaller learning rate for finetuning than for training on the source, implying that \( \alpha \) is even smaller than \( \beta \). 
Thus, we expect the constants $\sqrt{\frac{\alpha^2}{\beta}+\frac{(1-\alpha)^2}{1-\beta}}$ and $(1-\alpha)$ to both be close to $1$. 
Although the VC-dimension $d$ can be vacuously large for neural networks (a well-known defect in statistical learning theory), having a large enough sample size $m$ can generally make the first term comparable to the next two terms.
For the raw features $\phi_{\rm tabular}(X) = X$, we have observed empirically that 
\( Y|X \)-shifts are salient, implying that the third term (related to \( Y|X \)-shifts) dominates the second term (related to \( X \)-shifts), and it can also dominate the first term when $Y|X$ shifts are particular significant. 
In comparison, 
we conjecture that LLM embeddings $\phi_{\rm LLM}$ can reduce 
the gap between $\mathbb{E}_P[\mathbb{I}(Y \neq h(z) \mid \phi_{\rm LLM}(X) = z]$
and $\mathbb{E}_Q[\mathbb{I}(Y \neq h(z)) \mid \phi_{\rm LLM}(X) = z]$
by \emph{incorporating prior knowledge} encoded during LLM pre-training.\
Since $\epsilon_P(h)  = \int \mathbb{E}_P[\mathbb{I}(Y \neq h(z)) \mid \phi(X) = z]~\mathrm{d}P_{\phi(X)}(z)$ (and similarly for $Q$), we thus expect the third term to be 
much smaller than using raw features $\phi_{\rm tabular}(X)$. 
This suggests that when using LLM embeddings $\phi_{\rm LLM}$, the generalization bound can be smaller than that of raw features $\phi_{\rm tabular}(X)$, especially when $Y|X$ shift is more significant under raw features $\phi_{\rm tabular}(X) = X$; recall~\Cref{fig:curve}.

We hope our empirical findings spur future theoretical investigations into the foundations of LLM-based target adaptation.

% \section{Conclusion} 
% In this work, through a comprehensive and systematic study, we found that LLM embeddings alone offer inconsistent improvements in robustness under $Y|X$-shifts. However, models trained on these embeddings can effectively adapt to the target domain with only a few target samples. Our findings introduce a new approach to incorporating LLMs into tabular prediction, and future research should explore the theoretical interpretations in greater depth.

%% \paragraph{Acknowledgement} We are indebted to Lin Fan and Daniel Russo for
%% constructive feedback and helpful discussions. This research was partially
%% supported by the Digital Futures Initiative.

% Acknowledgments---Will not appear in anonymized version

%% ========================== Bibliography =========================  = %%

\newpage
\bibliographystyle{abbrvnat}

\ifdefined\useorstyle
\setlength{\bibsep}{.0em}
\else
\setlength{\bibsep}{.7em}
\fi

\bibliography{main.bbl}

\begin{thebibliography}{65}
\providecommand{\natexlab}[1]{#1}
\providecommand{\url}[1]{\texttt{#1}}
\expandafter\ifx\csname urlstyle\endcsname\relax
  \providecommand{\doi}[1]{doi: #1}\else
  \providecommand{\doi}{doi: \begingroup \urlstyle{rm}\Url}\fi

\bibitem[Amorim et~al.(2018)Amorim, Can{\c{c}}ado, and Veloso]{AmorimCaVe18}
E.~Amorim, M.~Can{\c{c}}ado, and A.~Veloso.
\newblock Automated essay scoring in the presence of biased ratings.
\newblock In \emph{Association for Computational Linguistics (ACL)}, pages 229--237, 2018.

\bibitem[Arik and Pfister(2021)]{arik2021tabnet}
S.~{\"O}. Arik and T.~Pfister.
\newblock Tabnet: Attentive interpretable tabular learning.
\newblock In \emph{Proceedings of the AAAI conference on artificial intelligence}, volume~35, pages 6679--6687, 2021.

\bibitem[Arjovsky et~al.(2019)Arjovsky, Bottou, Gulrajani, and Lopez-Paz]{ArjovskyBoGuLo19}
M.~Arjovsky, L.~Bottou, I.~Gulrajani, and D.~Lopez-Paz.
\newblock Invariant risk minimization.
\newblock \emph{arXiv:1907.02893 [stat.ML]}, 2019.

\bibitem[Bandi et~al.(2018)Bandi, Geessink, Manson, Van~Dijk, Balkenhol, Hermsen, Bejnordi, Lee, Paeng, Zhong, et~al.]{BandiEtAl18}
P.~Bandi, O.~Geessink, Q.~Manson, M.~Van~Dijk, M.~Balkenhol, M.~Hermsen, B.~E. Bejnordi, B.~Lee, K.~Paeng, A.~Zhong, et~al.
\newblock From detection of individual metastases to classification of lymph node status at the patient level: the camelyon17 challenge.
\newblock \emph{IEEE transactions on medical imaging}, 38\penalty0 (2):\penalty0 550--560, 2018.

\bibitem[Ben-David et~al.(2010)Ben-David, Blitzer, Crammer, Kulesza, Pereira, and Vaughan]{ben2010theory}
S.~Ben-David, J.~Blitzer, K.~Crammer, A.~Kulesza, F.~Pereira, and J.~W. Vaughan.
\newblock A theory of learning from different domains.
\newblock \emph{Machine learning}, 79:\penalty0 151--175, 2010.

\bibitem[Blanchet et~al.(2017)Blanchet, Kang, Zhang, and Murthy]{BlanchetKaZhMu17}
J.~Blanchet, Y.~Kang, F.~Zhang, and K.~Murthy.
\newblock Data-driven optimal transport cost selection for distributionally robust optimizatio.
\newblock \emph{arXiv:1705.07152 [stat.ML]}, 2017.

\bibitem[Blanchet et~al.(2018)Blanchet, Murthy, and Zhang]{BlanchetMuZh18}
J.~Blanchet, K.~Murthy, and F.~Zhang.
\newblock Optimal transport based distributionally robust optimization: Structural properties and iterative schemes.
\newblock \emph{arXiv:1810.02403 [stat.ML]}, 2018.

\bibitem[Blanchet et~al.(2019{\natexlab{a}})Blanchet, Kang, and Murthy]{BlanchetKaMu19}
J.~Blanchet, Y.~Kang, and K.~Murthy.
\newblock Robust {W}asserstein profile inference and applications to machine learning.
\newblock \emph{Journal of Applied Probability}, 56\penalty0 (3):\penalty0 830--857, 2019{\natexlab{a}}.

\bibitem[Blanchet et~al.(2019{\natexlab{b}})Blanchet, Kang, Murthy, and Zhang]{blanchet2019data}
J.~Blanchet, Y.~Kang, K.~Murthy, and F.~Zhang.
\newblock Data-driven optimal transport cost selection for distributionally robust optimization.
\newblock In \emph{2019 winter simulation conference (WSC)}, pages 3740--3751. IEEE, 2019{\natexlab{b}}.

\bibitem[Blanchet et~al.(2023{\natexlab{a}})Blanchet, Kuhn, Li, and Taskesen]{BlanchetKuLiTRa23}
J.~Blanchet, D.~Kuhn, J.~Li, and B.~Taskesen.
\newblock Unifying distributionally robust optimization via optimal transport theory.
\newblock \emph{arXiv:2308.05414 [math.OC]}, 2023{\natexlab{a}}.

\bibitem[Blanchet et~al.(2023{\natexlab{b}})Blanchet, Kuhn, Li, and Taskesen]{blanchet2023unifying}
J.~Blanchet, D.~Kuhn, J.~Li, and B.~Taskesen.
\newblock Unifying distributionally robust optimization via optimal transport theory.
\newblock \emph{arXiv preprint arXiv:2308.05414}, 2023{\natexlab{b}}.

\bibitem[Brown et~al.(2020)Brown, Mann, Ryder, Subbiah, Kaplan, Dhariwal, Neelakantan, Shyam, Sastry, and Askell]{BrownEtAl20}
T.~B. Brown, B.~Mann, N.~Ryder, M.~Subbiah, J.~Kaplan, P.~Dhariwal, A.~Neelakantan, P.~Shyam, G.~Sastry, and A.~Askell.
\newblock Language models are few-shot learners.
\newblock In \emph{Advances in Neural Information Processing Systems 33}, 2020.

\bibitem[Cai et~al.(2023)Cai, Namkoong, and Yadlowsky]{CaiNaYa23}
T.~T. Cai, H.~Namkoong, and S.~Yadlowsky.
\newblock Diagnosing model performance under distribution shift.
\newblock \emph{arXiv preprint arXiv:2303.02011}, 2023.

\bibitem[Chen et~al.(2023)Chen, Zhang, Song, Shan, and Liu]{chen2023improved}
L.~Chen, Y.~Zhang, Y.~Song, Y.~Shan, and L.~Liu.
\newblock Improved test-time adaptation for domain generalization.
\newblock In \emph{Proceedings of the IEEE/CVF Conference on Computer Vision and Pattern Recognition}, pages 24172--24182, 2023.

\bibitem[Chen and Guestrin(2016)]{chen2016xgboost}
T.~Chen and C.~Guestrin.
\newblock {{XGBoost}}: {{A Scalable Tree Boosting System}}.
\newblock In \emph{ACM SIGKDD International Conference on Knowledge Discovery}, pages 785--794. {ACM}, 2016.

\bibitem[Ding et~al.(2021{\natexlab{a}})Ding, Hardt, Miller, and Schmidt]{DingHaMoSc21}
F.~Ding, M.~Hardt, J.~Miller, and L.~Schmidt.
\newblock Retiring adult: New datasets for fair machine learning.
\newblock \emph{Advances in Neural Information Processing Systems 34}, 34, 2021{\natexlab{a}}.

\bibitem[Ding et~al.(2021{\natexlab{b}})Ding, Hardt, Miller, and Schmidt]{ding2021retiring}
F.~Ding, M.~Hardt, J.~Miller, and L.~Schmidt.
\newblock Retiring adult: New datasets for fair machine learning.
\newblock \emph{Advances in neural information processing systems}, 34:\penalty0 6478--6490, 2021{\natexlab{b}}.

\bibitem[Dinh et~al.(2022)Dinh, Zeng, Zhang, Lin, Gira, Rajput, Sohn, Papailiopoulos, and Lee]{DinhEtAl22}
T.~Dinh, Y.~Zeng, R.~Zhang, Z.~Lin, M.~Gira, S.~Rajput, J.-y. Sohn, D.~Papailiopoulos, and K.~Lee.
\newblock Lift: Language-interfaced fine-tuning for non-language machine learning tasks.
\newblock \emph{Advances in Neural Information Processing Systems}, 35:\penalty0 11763--11784, 2022.

\bibitem[Duchi and Namkoong(2019)]{DuchiNa19}
J.~C. Duchi and H.~Namkoong.
\newblock Variance-based regularization with convex objectives.
\newblock \emph{Journal of Machine Learning Research}, 20\penalty0 (68):\penalty0 1--55, 2019.

\bibitem[Duchi and Namkoong(2021)]{DuchiNa21}
J.~C. Duchi and H.~Namkoong.
\newblock Learning models with uniform performance via distributionally robust optimization.
\newblock \emph{Annals of Statistics}, 49\penalty0 (3):\penalty0 1378--1406, 2021.

\bibitem[Duchi et~al.(2021)Duchi, Glynn, and Namkoong]{DuchiGlNa21}
J.~C. Duchi, P.~W. Glynn, and H.~Namkoong.
\newblock Statistics of robust optimization: A generalized empirical likelihood approach.
\newblock \emph{Mathematics of Operations Research}, 46:\penalty0 946--969, 2021.

\bibitem[Friedman(2001)]{friedman2001greedy}
J.~H. Friedman.
\newblock Greedy function approximation: a gradient boosting machine.
\newblock \emph{Annals of statistics}, pages 1189--1232, 2001.

\bibitem[Friedman(2002)]{friedman2002stochastic}
J.~H. Friedman.
\newblock Stochastic gradient boosting.
\newblock \emph{Computational statistics \& data analysis}, 38\penalty0 (4):\penalty0 367--378, 2002.

\bibitem[Gardner et~al.(2022)Gardner, Popovic, and Schmidt]{gardner2022subgroup}
J.~Gardner, Z.~Popovic, and L.~Schmidt.
\newblock Subgroup robustness grows on trees: An empirical baseline investigation.
\newblock \emph{Advances in Neural Information Processing Systems}, 35:\penalty0 9939--9954, 2022.

\bibitem[Gorishniy et~al.(2021)Gorishniy, Rubachev, Khrulkov, and Babenko]{gorishniy2021revisiting}
Y.~Gorishniy, I.~Rubachev, V.~Khrulkov, and A.~Babenko.
\newblock Revisiting deep learning models for tabular data.
\newblock \emph{Advances in Neural Information Processing Systems}, 34:\penalty0 18932--18943, 2021.

\bibitem[Hand(2006)]{Hand06}
D.~J. Hand.
\newblock Classifier technology and the illusion of progress.
\newblock \emph{Statistical Science}, 21\penalty0 (1):\penalty0 1--14, 2006.

\bibitem[Hegselmann et~al.(2023)Hegselmann, Buendia, Lang, Agrawal, Jiang, and Sontag]{hegselmann2023tabllm}
S.~Hegselmann, A.~Buendia, H.~Lang, M.~Agrawal, X.~Jiang, and D.~Sontag.
\newblock Tabllm: Few-shot classification of tabular data with large language models.
\newblock In \emph{International Conference on Artificial Intelligence and Statistics}, pages 5549--5581. PMLR, 2023.

\bibitem[Hollmann et~al.(2022)Hollmann, M{\"u}ller, Eggensperger, and Hutter]{HollmannMuEgHu22}
N.~Hollmann, S.~M{\"u}ller, K.~Eggensperger, and F.~Hutter.
\newblock Tabpfn: A transformer that solves small tabular classification problems in a second.
\newblock \emph{arXiv preprint arXiv:2207.01848}, 2022.

\bibitem[Hu et~al.()Hu, Wallis, Allen-Zhu, Li, Wang, Wang, Chen, et~al.]{hulora}
E.~J. Hu, P.~Wallis, Z.~Allen-Zhu, Y.~Li, S.~Wang, L.~Wang, W.~Chen, et~al.
\newblock Lora: Low-rank adaptation of large language models.
\newblock In \emph{International Conference on Learning Representations}.

\bibitem[Hu and Hong(2013)]{hu2013kullback}
Z.~Hu and L.~J. Hong.
\newblock Kullback-leibler divergence constrained distributionally robust optimization.
\newblock \emph{Available at Optimization Online}, 1\penalty0 (2):\penalty0 9, 2013.

\bibitem[Huang et~al.(2020)Huang, Khetan, Cvitkovic, and Karnin]{huang2020tabtransformer}
X.~Huang, A.~Khetan, M.~Cvitkovic, and Z.~Karnin.
\newblock Tabtransformer: Tabular data modeling using contextual embeddings.
\newblock \emph{arXiv preprint arXiv:2012.06678}, 2020.

\bibitem[Iwasawa and Matsuo(2021)]{NEURIPS2021_1415fe9f}
Y.~Iwasawa and Y.~Matsuo.
\newblock Test-time classifier adjustment module for model-agnostic domain generalization.
\newblock In M.~Ranzato, A.~Beygelzimer, Y.~Dauphin, P.~Liang, and J.~W. Vaughan, editors, \emph{Advances in Neural Information Processing Systems}, volume~34, pages 2427--2440. Curran Associates, Inc., 2021.

\bibitem[Kadra et~al.(2021)Kadra, Lindauer, Hutter, and Grabocka]{kadra2021well}
A.~Kadra, M.~Lindauer, F.~Hutter, and J.~Grabocka.
\newblock Well-tuned simple nets excel on tabular datasets.
\newblock \emph{Advances in neural information processing systems}, 34:\penalty0 23928--23941, 2021.

\bibitem[Katzir et~al.(2020)Katzir, Elidan, and El-Yaniv]{katzir2020net}
L.~Katzir, G.~Elidan, and R.~El-Yaniv.
\newblock Net-dnf: Effective deep modeling of tabular data.
\newblock In \emph{International conference on learning representations}, 2020.

\bibitem[Ke et~al.(2017)Ke, Meng, Finley, Wang, Chen, Ma, Ye, and Liu]{ke2017lightgbm}
G.~Ke, Q.~Meng, T.~Finley, T.~Wang, W.~Chen, W.~Ma, Q.~Ye, and T.-Y. Liu.
\newblock Lightgbm: A highly efficient gradient boosting decision tree.
\newblock \emph{Advances in neural information processing systems}, 30, 2017.

\bibitem[Koyama and Yamaguchi(2020)]{koyama2020invariance}
M.~Koyama and S.~Yamaguchi.
\newblock When is invariance useful in an out-of-distribution generalization problem?
\newblock \emph{arXiv preprint arXiv:2008.01883}, 2020.

\bibitem[Li et~al.(2017)Li, Yang, Song, and Hospedales]{li2017deeper}
D.~Li, Y.~Yang, Y.-Z. Song, and T.~M. Hospedales.
\newblock Deeper, broader and artier domain generalization.
\newblock In \emph{Proceedings of the IEEE international conference on computer vision}, pages 5542--5550, 2017.

\bibitem[Li et~al.(2022)Li, Cai, and Li]{li2022transfer}
S.~Li, T.~T. Cai, and H.~Li.
\newblock Transfer learning for high-dimensional linear regression: Prediction, estimation and minimax optimality.
\newblock \emph{Journal of the Royal Statistical Society Series B: Statistical Methodology}, 84\penalty0 (1):\penalty0 149--173, 2022.

\bibitem[Li and Liang(2021)]{li2021prefix}
X.~L. Li and P.~Liang.
\newblock Prefix-tuning: Optimizing continuous prompts for generation.
\newblock In \emph{Proceedings of the 59th Annual Meeting of the Association for Computational Linguistics and the 11th International Joint Conference on Natural Language Processing (Volume 1: Long Papers)}, pages 4582--4597, 2021.

\bibitem[Liang et~al.(2023)Liang, He, and Tan]{liang2023ttasurvey}
J.~Liang, R.~He, and T.~Tan.
\newblock A comprehensive survey on test-time adaptation under distribution shifts.
\newblock \emph{International Journal Of Computer Vision}, 2023.

\bibitem[Liu et~al.(2023)Liu, Wang, Cui, and Namkoong]{LiuWaCuNa23}
J.~Liu, T.~Wang, P.~Cui, and H.~Namkoong.
\newblock On the need for a language describing distribution shifts: Illustrations on tabular datasets.
\newblock In \emph{Advances in Neural Information Processing Systems 36}, 2023.

\bibitem[McElfresh et~al.(2024)McElfresh, Khandagale, Valverde, Prasad~C, Ramakrishnan, Goldblum, and White]{mcelfresh2024neural}
D.~McElfresh, S.~Khandagale, J.~Valverde, V.~Prasad~C, G.~Ramakrishnan, M.~Goldblum, and C.~White.
\newblock When do neural nets outperform boosted trees on tabular data?
\newblock \emph{Advances in Neural Information Processing Systems}, 36, 2024.

\bibitem[Miller et~al.(2021)Miller, Taori, Raghunathan, Sagawa, Koh, Shankar, Liang, Carmon, and Schmidt]{MillerTaRaSaKoShLiCaSc21}
J.~Miller, R.~Taori, A.~Raghunathan, S.~Sagawa, P.~W. Koh, V.~Shankar, P.~Liang, Y.~Carmon, and L.~Schmidt.
\newblock Accuracy on the line: on the strong correlation between out-of-distribution and in-distribution generalization.
\newblock In \emph{Proceedings of the 38th International Conference on Machine Learning}, 2021.

\bibitem[Natekin and Knoll(2013)]{natekin2013gradient}
A.~Natekin and A.~Knoll.
\newblock Gradient boosting machines, a tutorial.
\newblock \emph{Frontiers in neurorobotics}, 7:\penalty0 21, 2013.

\bibitem[{OpenAI}(2023)]{OpenAI23}
{OpenAI}.
\newblock {GPT}-4 technical report.
\newblock \emph{arXiv preprint arXiv:2303.08774}, 2023.

\bibitem[{OpenAI}(2024)]{OpenAI24}
{OpenAI}.
\newblock {GPT}-4o mini: advancing cost-efficient intelligence, 2024.
\newblock URL \url{https://openai.com/index/gpt-4o-mini-advancing-cost-efficient-intelligence/}.
\newblock Accessed October 2024.

\bibitem[Pedregosa et~al.(2011)Pedregosa, Varoquaux, Gramfort, Michel, Thirion, Grisel, Blondel, Prettenhofer, Weiss, Dubourg, VanderPlas, Passos, Cournapeau, Brucher, Perrot, and Duchesnay]{sklearn}
F.~Pedregosa, G.~Varoquaux, A.~Gramfort, V.~Michel, B.~Thirion, O.~Grisel, M.~Blondel, P.~Prettenhofer, R.~Weiss, V.~Dubourg, J.~VanderPlas, A.~Passos, D.~Cournapeau, M.~Brucher, M.~Perrot, and E.~Duchesnay.
\newblock Scikit-learn: Machine learning in python.
\newblock \emph{J. Mach. Learn. Res.}, 12:\penalty0 2825--2830, 2011.

\bibitem[Peters et~al.(2016)Peters, B{\"u}hlmann, and Meinshausen]{peters2016causal}
J.~Peters, P.~B{\"u}hlmann, and N.~Meinshausen.
\newblock Causal inference by using invariant prediction: identification and confidence intervals.
\newblock \emph{Journal of the Royal Statistical Society Series B: Statistical Methodology}, 78\penalty0 (5):\penalty0 947--1012, 2016.

\bibitem[Recht et~al.(2019)Recht, Roelofs, Schmidt, and Shankar]{RechtRoScSh19}
B.~Recht, R.~Roelofs, L.~Schmidt, and V.~Shankar.
\newblock Do {I}mage{N}et classifiers generalize to {I}mage{N}et?
\newblock In \emph{Proceedings of the 36th International Conference on Machine Learning}, 2019.

\bibitem[Rockafellar et~al.(2000)Rockafellar, Uryasev, et~al.]{rockafellar2000optimization}
R.~T. Rockafellar, S.~Uryasev, et~al.
\newblock Optimization of conditional value-at-risk.
\newblock \emph{Journal of risk}, 2:\penalty0 21--42, 2000.

\bibitem[Rosenfeld et~al.(2021)Rosenfeld, Ravikumar, and Risteski]{RosenfeldRaRi20}
E.~Rosenfeld, P.~Ravikumar, and A.~Risteski.
\newblock The risks of invariant risk minimization.
\newblock In \emph{Proceedings of the Ninth International Conference on Learning Representations}, 2021.

\bibitem[Schick and Sch{\"u}tze(2020)]{SchickSc20}
T.~Schick and H.~Sch{\"u}tze.
\newblock Exploiting cloze questions for few shot text classification and natural language inference.
\newblock \emph{arXiv preprint arXiv:2001.07676}, 2020.

\bibitem[Shankar et~al.(2019)Shankar, Dave, Roelofs, Ramanan, Recht, and Schmidt]{ShankarDaRoRaReSc19}
V.~Shankar, A.~Dave, R.~Roelofs, D.~Ramanan, B.~Recht, and L.~Schmidt.
\newblock Do image classifiers generalize across time?
\newblock \emph{arXiv:1906.02168 [cs.LG]}, 2019.

\bibitem[Shin et~al.(2020)Shin, Razeghi, Logan~IV, Wallace, and Singh]{ShinEtAl20}
T.~Shin, Y.~Razeghi, R.~L. Logan~IV, E.~Wallace, and S.~Singh.
\newblock Autoprompt: Eliciting knowledge from language models with automatically generated prompts.
\newblock \emph{arXiv preprint arXiv:2010.15980}, 2020.

\bibitem[Shwartz-Ziv and Armon(2022)]{shwartz2022tabular}
R.~Shwartz-Ziv and A.~Armon.
\newblock Tabular data: Deep learning is not all you need.
\newblock \emph{Information Fusion}, 81:\penalty0 84--90, 2022.

\bibitem[Slack and Singh(2023)]{SlackSi23}
D.~Slack and S.~Singh.
\newblock Tablet: Learning from instructions for tabular data.
\newblock \emph{arXiv preprint arXiv:2304.13188}, 2023.

\bibitem[Tian and Feng(2023)]{tian2023transfer}
Y.~Tian and Y.~Feng.
\newblock Transfer learning under high-dimensional generalized linear models.
\newblock \emph{Journal of the American Statistical Association}, 118\penalty0 (544):\penalty0 2684--2697, 2023.

\bibitem[Wang et~al.(2023{\natexlab{a}})Wang, Yang, Huang, Yang, Majumder, and Wei]{WangEtAl23}
L.~Wang, N.~Yang, X.~Huang, L.~Yang, R.~Majumder, and F.~Wei.
\newblock Improving text embeddings with large language models.
\newblock \emph{arXiv preprint arXiv:2401.00368}, 2023{\natexlab{a}}.

\bibitem[Wang et~al.(2023{\natexlab{b}})Wang, Wang, and Sun]{WangWaSu23}
R.~Wang, Z.~Wang, and J.~Sun.
\newblock Unipredict: Large language models are universal tabular predictors.
\newblock \emph{arXiv preprint arXiv:2310.03266}, 2023{\natexlab{b}}.

\bibitem[Wen et~al.(2024)Wen, Zhang, Zheng, Xu, and Bian]{WenEtAl24}
X.~Wen, H.~Zhang, S.~Zheng, W.~Xu, and J.~Bian.
\newblock From supervised to generative: A novel paradigm for tabular deep learning with large language models.
\newblock In \emph{Proceedings of the 30th ACM SIGKDD Conference on Knowledge Discovery and Data Mining}, pages 3323--3333, 2024.

\bibitem[Wong et~al.(2021)Wong, Otles, Donnelly, Krumm, McCullough, DeTroyer-Cooley, Pestrue, Phillips, Konye, Penoza, et~al.]{WongEtAl21}
A.~Wong, E.~Otles, J.~P. Donnelly, A.~Krumm, J.~McCullough, O.~DeTroyer-Cooley, J.~Pestrue, M.~Phillips, J.~Konye, C.~Penoza, et~al.
\newblock External validation of a widely implemented proprietary sepsis prediction model in hospitalized patients.
\newblock \emph{JAMA Internal Medicine}, 181\penalty0 (8):\penalty0 1065--1070, 2021.

\bibitem[Yan et~al.(2024)Yan, Zheng, Xu, Zhu, Chen, Sun, Wu, and Chen]{YanEtAl24}
J.~Yan, B.~Zheng, H.~Xu, Y.~Zhu, D.~Chen, J.~Sun, J.~Wu, and J.~Chen.
\newblock Making pre-trained language models great on tabular prediction.
\newblock \emph{arXiv preprint arXiv:2403.01841}, 2024.

\bibitem[Yang et~al.(2024)Yang, Wang, Sen, Li, and Liu]{yang2024unleashing}
Y.~Yang, Y.~Wang, S.~Sen, L.~Li, and Q.~Liu.
\newblock Unleashing the potential of large language models for predictive tabular tasks in data science.
\newblock \emph{arXiv preprint arXiv:2403.20208}, 2024.

\bibitem[Zhou et~al.(2022)Zhou, Liu, Qiao, Xiang, and Loy]{zhou2022domain}
K.~Zhou, Z.~Liu, Y.~Qiao, T.~Xiang, and C.~C. Loy.
\newblock Domain generalization: A survey.
\newblock \emph{IEEE Transactions on Pattern Analysis and Machine Intelligence}, 45\penalty0 (4):\penalty0 4396--4415, 2022.

\bibitem[Zhuang et~al.(2020)Zhuang, Qi, Duan, Xi, Zhu, Zhu, Xiong, and He]{zhuang2020comprehensive}
F.~Zhuang, Z.~Qi, K.~Duan, D.~Xi, Y.~Zhu, H.~Zhu, H.~Xiong, and Q.~He.
\newblock A comprehensive survey on transfer learning.
\newblock \emph{Proceedings of the IEEE}, 109\penalty0 (1):\penalty0 43--76, 2020.

\end{thebibliography}

\ifdefined\useorstyle

\ECSwitch

%\ECDisclaimer
%%%%%%%%%%%%%%%%%%%%%%%%%%%%%%%%%%%%%%%%%%%%%%%%%%%%%%%%%%

%%% Main head for the e-companion
\ECHead{Appendix}

\else
\appendix
\newpage
\part*{Appendices} % Start the appendix part
%\addcontentsline{toc}{part}{Appendices} % Add the appendices to the main ToC
%\thepart{} % Start the document part
%\parttoc 

\section{Model Training Details}
\label{sec: model training details}

In this section, we outline the serialization scheme, the generation of additional domain information, our model architecture, and the hyperparameters used for training and target adaptation.

\subsection{Serialization Scheme}

As discussed in Section~\ref{section: llm embedding}, serializing a row of tabular data, such as $X$, requires two components: a task description and a description of the data.

\paragraph{Task description} For the ACS Income, ACS Mobility, and ACS Public Coverage datasets, we consider the same binary classification task as 
described in~\citet{{DingHaMoSc21}}. 
We adopt concise task descriptions as suggested by~\cite{hegselmann2023tabllm}, as follows:
\begin{itemize}
    \item \texttt{ACS Income}: Classify whether US working adults' yearly income is above \$50000 in 2018.
    \item \texttt{ACS Mobility}: Classify whether a young adult moved addresses in the last year.
    \item \texttt{ACS Public Coverage} : Classify whether a low-income individual, not eligible for Medicare, has coverage from public health insurance.
\end{itemize}

\paragraph{Description of the data}
\label{sec: appendix description of the data}

For all three ACS datasets, we utilize features as shown in~\citep[Appendix B]{DingHaMoSc21}, along with the domain name (state) to characterize the data. 
Specifically, for data from the source state, we apply the source domain name, and for data from the target state, we use the target domain name.

As illustrated in Figure~\ref{fig: method overview} and recommended by~\cite{hegselmann2023tabllm}, we adopt a straightforward text template: "The \texttt{feature name} is \texttt{value}." However, for certain features with less common or more complex feature names, we opt for a template as ``The person (appropriate verb) \texttt{value}" to more clearly convey the data description. For example, for the feature ``Gave birth to child within the past 12 months" with the value ``No", the serialization would be ``The person did not give birth to a child within the past 12 months." The features that use this alternative template are as follows:
\begin{itemize}
    \item \texttt{ACS Income}: class of worker;
    
    \item \texttt{ACS Mobility}: class of worker, disability, employment status of parents, citizenship, military service, hearing difficulty, vision difficulty, cognitive difficulty, grandparent living with grandchildren, employment status;
    
    \item \texttt{ACS Public Coverage}: disability, employment status of parents, citizenship, mobility, military service, hearing difficulty, vision difficulty, cognitive difficulty,  gave birth to child within the past 12 months, employment status.
\end{itemize}
For features without associated values, we omit them during serialization.

\subsection{Additional domain information}
\label{sec: appendix additional domain info}
As shown in Section~\ref{section: additional domain info}, we study three sources of domain information: Wikipedia, GPT-4, and labeled target samples. As we use the 
\texttt{e5-mistral-7b-instruct} to generate an LLM embedding for the domain information, we need to specify both the ``Instruct" and ``Query" components.

For the ``Instruct" part, we apply the same task description as outlined in Section~\ref{sec: appendix description of the data}. For the ``Query" part, we utilize various descriptions of the additional domain information:
\begin{itemize}
\item Wikipedia: For all three datasets, we use the ``Economy" section of each state's Wikipedia page as the additional domain information.
\item GPT-4: For each state, we pose the following question to GPT-4, and use its response as the additional domain:
\begin{itemize}
\item \texttt{ACS Income}: ``We aim to develop a classifier to determine whether U.S. individuals earned over \$50000 in 2018, using features such as age, sex, educational attainment, race, class of worker, marital status, occupation, and hours worked per week over the past 12 months. Given the unique economic and demographic profile of \texttt{state\_name}, how might these factors influence income levels differently compared to other U.S. states? Please provide a 2000-word summary detailing these differences."

\item \texttt{ACS Mobility}: ``We aim to develop a classifier to determine whether a young adult moved addresses in the last year, using features such as age, sex, educational attainment, race, class of worker, marital status, occupation, total income, and hours worked per week over the past 12 months. Given the unique economic and demographic profile of \texttt{state\_name}, how might these factors influence mobility levels differently compared to other U.S. states? Please provide a 2000-word summary detailing these differences."

\item \texttt{ACS Public Coverage}: ``We aim to develop a classifier to determine whether a low-income individual, not eligible for Medicare, has coverage from public health insurance, using features such as age, sex, educational attainment, race, disability, marital status, occupation, citizenship status, mobility status, military service, nativity, total income, and employment status. Given the unique economic and demographic profile of \texttt{state\_name}, how might these factors influence public coverage levels differently compared to other U.S. states? Please provide a 2000-word summary detailing these differences."
\end{itemize}

\item labeled target samples: We use the following template for 32 labeled target samples:
\begin{flushleft}
    \texttt{Here are some examples of the data: \textbackslash n} \\
    \texttt{description of one target sample  \textbackslash n} \\
    \texttt{Answer: (Yes or No). \textbackslash n  \textbackslash n}\\
    \texttt{description of another target sample  \textbackslash n}\\
    \texttt{Answer: (Yes or No). \textbackslash n  \textbackslash n}\\ ...
\end{flushleft}
We use the same data descriptions as in Section~\ref{sec: appendix description of the data}, with the exception that the state/domain name is omitted, as it is represented by the labeled target samples provided.
\end{itemize}

\subsection{Model Architecture}\label{sec:appendix model arch}

We detail the baselines used in our paper.

\paragraph{Fully-connected Neural Networks (NN)}
Given the varying input dimensions for Tabular features, LLM embeddings, and LLM embeddings with additional domain information, we employ three neural networks with similar architectures. We then train these networks and conduct target adaptation  using Empirical Risk Minimization (ERM).

For all three datasets using Tabular features, we use a hidden layer with output dimension being 
$\texttt{hidden layer dim}$ as a hyperparameter, a dropuput layer with \texttt{dropout ratio} being a hyperparameter, ReLu activation. We then have another hidden layer with input and output dimension both being $\texttt{hidden layer dim}$, and then softmax layer with dimension 2 as the output.

For datasets using Tabular features, the network includes a hidden layer where the output dimension is set by the hyperparameter \texttt{hidden layer dim}. It is followed by a dropout layer (\texttt{dropout ratio} as a hyperparameter), ReLU activation, another hidden layer the input and output dimensions are both equal to 
\texttt{hidden layer dim}. The network concludes with a softmax layer with an output dimension of 2.

For datasets using only LLM embeddings, the input dimension is 4096, which is the output dimension of \texttt{e5-mistral-7b-instruct}. The network consists of three hidden layers with fixed dimensions, where the input and output dimensions are (4096, 1024), (1024, 256), and (256, 128), respectively. Each layer uses ReLU activation. Next, there's a hidden layer with an input dimension of 128 and an output dimension set by the 
\texttt{hidden layer dim} hyperparameter. This is followed by a dropout layer (with 
\texttt{dropout ratio} as a hyperparameter), ReLU activation, and another hidden layer where both the input and output dimensions are 
\texttt{hidden layer dim}. A ReLU activation follows this final hidden layer, and the network concludes with a softmax layer that outputs a dimension of 2.

For datasets using LLM embeddings concatenated with additional domain information, the input dimension is 8192. We slightly update the first three hidden layers, with input and out dimensions being  (8192, 2048), (2048, 512), and (512, 128), respectively. All other neural network structure and hyperparameters are the same as NNs with LLM embeddings only. 

When applying low-rank adaptation (LoRA) for target adaptation, we introduce a low-rank adaptation layer to each linear layer by incorporating two smaller matrices, $A$ and $B$, both with a rank of 16. Specifically, matrix $A$ has dimensions corresponding to the input size and the rank, while matrix $B$ has dimensions corresponding to the rank and the output size. Matrix $A$ is initialized with a mean of 0 and a standard deviation of 0.02, whereas matrix $B$ is initialized with zeros. These matrices are then multiplied together and added to the original weight matrix.

\paragraph{Tree Ensemble Models}
\citet{gardner2022subgroup} show that several tree-based methods are competitive on tabular datasets.
And gradient-boosted trees (e.g., XGB~\citep{chen2016xgboost}, LGBM~\citep{ke2017lightgbm}, GBM~\citep{natekin2013gradient}) are widely considered as the state-of-the-art methods on tabular data.
Therefore, we compare XGB, LGBM, and GBM in this work.
For GBM, we use the standard implementations in \texttt{scikit-learn}~\citep{sklearn}.
For XGB and LGBM, we use the standard implementations in the \texttt{xgboost} package\footnote{\url{https://pypi.org/project/xgboost/}} and the \texttt{lightgbm} package\footnote{\url{https://pypi.org/project/lightgbm/}}.
All these methods are trained on CPUs.

\paragraph{DRO Methods}
Distributionally robust optimization (DRO) methods optimize the worst-case loss over an ambiguity set $\mathcal P$:
\begin{equation}
	\min_{f\in \mathcal F}\sup_{Q\in\mathcal P} \mathbb E_{Q}[\ell(f(X),Y)].
\end{equation}
The ambiguity set is typically chosen as a ``ball'' around the training distribution $P_{\text{tr}}$
\begin{equation}
	\mathcal P(d,\epsilon)=\{Q: d(Q,P_{\text{tr}})\leq \epsilon\},
\end{equation}
where $d(\cdot,\cdot)$ is a distance metric between probability measures and $\epsilon$ is the radius of set.
When $d$ is set as the generalized $f$-divergence (including CVaR) as:
\begin{equation}
	d(P, Q) = E_{Q}\bigg[f\big(\frac{dP}{dQ}\big)\bigg], 	
\end{equation}
for the KL-DRO method \citep{hu2013kullback}, we use $f(x) = x \log x - (x - 1)$;
for the $\chi^2$-DRO method \citep{DuchiNa19},  we use $f(x) = (x - 1)^2$;
for the CVaR-DRO problem~\citep{rockafellar2000optimization},
we use $f(x) = 0$ if $x \in [\frac{1}{\alpha}, \alpha]$ and $\infty$ otherwise, and $\alpha$ controls the worst-case ratio. 

For Wasserstein DRO~\citep{blanchet2019data}, we choose $d(\cdot,\cdot)$ as the Wasserstein distance.
Unified-DRO~\citep{blanchet2023unifying} combines Wasserstein distance and KL-divergence as $d(\cdot,\cdot)$, and we follow their initial Github codebases and hyperparameter selection when implementing these methods.

\begin{table}[t]
\vspace{-1cm}
\caption{Summary of methodologies: As baselines, we use basic models (LR, SVM, and NN), GBDTs (XGB, LGBM, and GBM), robust learning methods (KL-DRO, $\chi^2$-DRO, etc.), and recent advanced LLM in-context learning models (TabPFN and GPT-4-mini). 
For methods utilizing LLM embeddings, we consider different ways to incorporate domain information and different model architectures and finetuning techniques.}
\label{tab:overview-methods}
\vspace{0.1in}
\resizebox{\textwidth}{!}{\begin{tabular}{@{}lcccccc@{}}
\toprule
Name & Feature & Domain Info & Model & Adaptation & \# of HPs & Part of Fig.~\ref{fig: method overview}  \\ \midrule
LR & Tabular & - & LR & No Finetuning & 34 & -\\
SVM & Tabular & - & SVM & No Finetuning & 34 & -\\
GBDT & Tabular & - & XGB, LGBM, GBM & No Finetuning & 200 & - \\
KL-DRO & Tabular & - & SVM & No Finetuning & 138 & -\\
$\chi^2$-DRO & Tabular & - & SVM & No Finetuning & 138 &-\\
Wasserstein-DRO & Tabular & - & SVM & No Finetuning & 17 &-\\
Unified-DRO & Tabular & - & SVM & No Finetuning & 150 & -\\
CVaR-DRO & Tabular & - & NN & No Finetuning & 200 & - \\\midrule
\multirow{3}{*}{Tabular $|$ NN} & \multirow{3}{*}{Tabular} & \multirow{3}{*}{-} & \multirow{3}{*}{NN} & No Finetuning & 96 & - \\
 &  &  &  & Finetuning & 12 & - \\
 &  &  &  & LoRA & 15 & - \\\midrule
\begin{tabular}[l]{@{}l@{}}LLM In-Context \\
  Learning\end{tabular}  
& \multirow{2}{*}{Tabular} & - & TabPFN & No Finetuning & - & - \\
 &  & State Name & GPT-4-mini & No Finetuning & - & - \\\midrule
\multirow{3}{*}{LLM $|$ NN} & \multirow{3}{*}{\begin{tabular}[c]{@{}c@{}}LLM\\ Embeddings\end{tabular}} & \multirow{3}{*}{-} & \multirow{3}{*}{NN} & No Finetuning & 96 & E.1 \\
 &  &  &  & Finetuning & 12 & F.2 \\
 &  &  &  & LoRA & 15 & F.2 \\\midrule
\multirow{4}{*}{\begin{tabular}[l]{@{}l@{}}LLM \&  Wiki/\\ GPT-4 $|$ NN\end{tabular}} & \multirow{4}{*}{\begin{tabular}[c]{@{}c@{}}LLM\\ Embeddings\end{tabular}} & \multirow{4}{*}{\begin{tabular}[c]{@{}c@{}}  Wikipedia or  GPT-4\end{tabular}} & \multirow{4}{*}{\begin{tabular}[c]{@{}c@{}}NN 
\end{tabular}} & No Finetuning & 48 & E.2 \\
 &  &  &  & Finetuning & 12 & F.2 \\
 &  &  &  & LoRA & 15 & F.2\\ 
 &  &  &  & \begin{tabular}[c]{@{}c@{}}Prefix Tuning
\end{tabular} & 18 & F.3 \\\midrule
\begin{tabular}[l]{@{}l@{}}LLM \& In-Context\\ Domain Info $|$ NN\end{tabular} & \begin{tabular}[c]{@{}c@{}}LLM \\ Embeddings\end{tabular} & \begin{tabular}[c]{@{}c@{}}Labeled \\ Target Samples\end{tabular} & NN & \multicolumn{1}{c}{No Finetuning} & \multicolumn{1}{c}{96} & F.1 \\ \bottomrule
\end{tabular}}
\end{table}

\subsection{Hyperparameters for Training and Target Adaptation}
\label{sec: appendix hps}

For each algorithm, we maintain a grid of candidate hyperparameters as shown in 
Tables~\ref{table: training HPs} and~\ref{table: target adaptation HPs}.
and perform hyperparameter selection as described in Section~\ref{sec: testbed setup}. 
When the number of hyperparameter configurations exceeds 200, we randomly select 200 configurations to reduce computational cost and maintain fairness in the comparison across all algorithms.

\begin{table}[]
\caption{Training hyperparameter grids used in all experiments. 
$\diamond:$ for methods with the total grid size above 200, we randomly sample 200 configurations for fair comparisons.}
\label{table: training HPs}
\vspace{0.1in}
\resizebox{\textwidth}{!}{\begin{tabular}{@{}ccccccc@{}}
\toprule
Model & Feature & Domain Info & \# of HPs & Hyperparameter & Value Range 
\\ \midrule
\multirow{4}{*}{SVM} & \multirow{4}{*}{
\begin{tabular}[c]{@{}c@{}}
\textsf{Tabular}
\end{tabular}} & \multirow{4}{*}{-} &  \multirow{4}{*}{96} & C & {\footnotesize $\{ 1e^{-2}, 1e^{-1}, 1,1e^1,1e^2, 1e^3 \}$} \\
& & & & Kernel & $\{ \text{linear}, \text{RBF}\}$  \\
& & & & Loss & Squared Hinge \\
& & & & $\gamma$ & {\footnotesize $\{ 0.1, 0.3, 0.5, 1, 1.5, 2, \text{scale}, \text{auto} \}$} \\\midrule

 \multirow{3}{*}{LR} &  \multirow{3}{*}{\textsf{Tabular}} &  \multirow{3}{*}{-} & \multirow{3}{*}{23} & \multirow{3}{*}{$L_2$ penalty} & $\{1e^{-3}, 3e^{-3}, 5e^{-3}, 7e^{-3}, 1e^{-2},$ \\ 
& &  &  &  & $3e^{-2},5e^{-2},\dots, 1.3,1.7,5\}$\\
& &  &  &  & $1e^1,5e^1, 1e^2, 5e^2,1e^3,5e^3, 1e^4\}$\\\midrule

\multirow{4}{*}{Tabular $|$ NN} 
& \multirow{4}{*}{
\begin{tabular}[c]{@{}c@{}}
\textsf{Tabular}
\end{tabular}} 
& \multirow{4}{*}{-} 
& \multirow{4}{*}{96} 
& Learning Rate & $\{0.001, 0.003, 0.005, 0.01\}$  \\
& & & & Hidden Layer Dim &  $\{16, 32, 64, 128\}$   \\
& & & & Dropout Ratio &  $\{0, 0.1\}$ \\ 
& & & & Train Epoch & $\{50, 100, 200\}$\\\midrule

\multirow{5}{*}{GBM} & \multirow{5}{*}{\textsf{Tabular}} & \multirow{5}{*}{-}& \multirow{5}{*}{1680$^\diamond$} & Learning Rate & $\{ 1e^{-2}, 1e^{-1}, 5e^{-1} , 1 \}$ \\
& &  & & Num. Estimators & $\{ 32, 64, 128, 256\}$  \\
& & & & Max Depth & $\{ 2,4,8,16\}$ \\
& & & & Min. Child Samples & $\{ 1, 2, 4, 8\}$ \\ \midrule
 \multirow{5}{*}{LGBM} & \multirow{5}{*}{\textsf{Tabular}} & \multirow{5}{*}{-} & \multirow{5}{*}{1680$^\diamond$} & Learning Rate & $\{ 1e^{-2}, 1e^{-1}, 5e^{-1} , 1 \}$ \\
& & & & Num. Estimators & $\{ 64, 128, 256, 512 \}$  \\
& & & & $L_2$-reg. & $\{ 0, 1e^{-3}, 1e^{-2}, 1e^{-1}, 1 \}$ \\
& & & & Min. Child Samples & $\{ 1, 2, 4, 8, 16, 32, 64 \}$ \\
& & & & {\footnotesize Column Subsample Ratio (tree)} & $\{ 0.5, 0.8, 1. \}$ \\ \midrule
\multirow{7}{*}{XGB}& \multirow{7}{*}{\textsf{Tabular}}& \multirow{7}{*}{-} & \multirow{7}{*}{1944$^\diamond$} & Learning Rate & $\{ 0.1, 0.3, 1.0, 2.0\}$ \\
& & & & Min. Split Loss & $\{ 0, 0.1, 0.5 \}$  \\
& & & & Max. Depth & $\{ 4, 6, 8 \}$ \\
& & & & {\footnotesize Column Subsample Ratio (tree)} & $\{ 0.7, 0.9, 1 \}$ \\
& & & & {\footnotesize Column Subsample Ratio (level)} & $\{ 0.7, 0.9, 1 \}$ \\
& & & & Max. Bins & $\{ 128, 256, 512 \}$ \\ 
& & & & Growth Policy & {\footnotesize$\{ \textrm{Depthwise}, \textrm{Loss Guide} \}$} \\ \bottomrule

KL-DRO &\textsf{Tabular}& - & 117 & {\footnotesize Uncertainty Set Size $\epsilon$} & $\{1e^{-4}, \dots, 0.01, \dots, 0.99\}$\\ 
$\chi^2$-DRO &\textsf{Tabular}& - & 117 & {\footnotesize Uncertainty Set Size $\epsilon$} & $\{1e^{-4}, \dots, 0.01, \dots, 0.99\}$\\  
{\small Wasserstein-DRO} &\textsf{Tabular}& -& 138 & {\footnotesize Uncertainty Set Size $\epsilon$} & {\footnotesize $\{1e^{-4}, \dots, 0.01, \dots, 0.99, \dots, 3\}$}\\\midrule
\multirow{2}{*}{\small CVaR-DRO} & \multirow{2}{*}{\textsf{Tabular}} & \multirow{2}{*}{-} &\multirow{2}{*}{1620$^\diamond$} & Worst-case Ratio $\alpha$ & $\{0.01, 0.1, 0.2, 0.3, 0.5, 1.0\}$\\
& & & & Underlying Model Class & NN\\ \midrule

\multirow{3}{*}{Unified-DRO}&\multirow{3}{*}{\textsf{Tabular}} &\multirow{3}{*}{-} &\multirow{3}{*}{180} & Distance Type & $L_{\text{inf}}$\\
& & & & {\footnotesize Uncertainty Set Size $\epsilon$} & $\{1e^{-3}, \dots, 9e^{-1}\}$\\
& & & &$\theta_1$ & \tiny $\{1.001, 1.01, 1.1, 1.5, 2,3,5,10,50,100\}$\\
\midrule

\multirow{4}{*}{LLM $|$ NN} 
& \multirow{4}{*}{
\begin{tabular}[c]{@{}c@{}}
\textsf{LLM}
\end{tabular}} 
& \multirow{4}{*}{-} 
& \multirow{4}{*}{48} 
& Learning Rate & $\{0.001, 0.01\}$  \\
& & & & Hidden Layer Dim &  $\{32, 64, 128\}$   \\
& & & & Dropout Ratio &  $\{0, 0.1\}$ \\ 
& & & & Train Epoch & $\{100, 200, 300, 500\}$\\\midrule
\multirow{4}{*}{
\begin{tabular}[c]{@{}c@{}}
LLM \& In-Context Domain \\Info/Wiki/GPT-4 $|$  NN
\end{tabular}} 
& \multirow{4}{*}{
\begin{tabular}[c]{@{}c@{}}
\textsf{LLM}
\end{tabular}} 
& \multirow{4}{*}{
\begin{tabular}[c]{@{}c@{}}
Labeled
Target  Samples\\
or Wikipedia \\or GPT-4
\end{tabular}} 
& \multirow{4}{*}{48} 
& Learning Rate & $\{0.001, 0.01\}$  \\
& & & & Hidden Layer Dim &  $\{32, 64, 128\}$   \\
& & & & Dropout Ratio &  $\{0, 0.1\}$ \\ 
& & & & Train Epoch & $\{100, 200, 300, 500\}$\\
\bottomrule
\end{tabular}}
\end{table}

\begin{table}[]
\caption{Target adaptation hyperparameter grids used in all experiments.}
\label{table: target adaptation HPs}
\vspace{0.1in}
\resizebox{\textwidth}{!}{\begin{tabular}{@{}ccccccc@{}}
\toprule
Model & Target Adaptation Method
& \# of HPs & Hyperparameter & Value Range 
\\ \midrule
\multirow{2}{*}{
\begin{tabular}[c]{@{}c@{}}
Tabular $|$ NN 
or LLM $|$ NN \\
or LLM \& Wiki/GPT-4 $|$ NN
\end{tabular}} 
& \multirow{2}{*}{
\begin{tabular}[c]{@{}c@{}}
Finetuning
\end{tabular}} 
& \multirow{2}{*}{-} 
& \multirow{2}{*}{12} 
& Learning Rate & $\{10^{-7}, 10^{-6}, 10^{-5}, 10^{-4}\}$  \\
& & & & Target adaptation Epoch & $\{25, 50, 100\}$ \\\midrule
\multirow{2}{*}{
\begin{tabular}[c]{@{}c@{}}
Tabular $|$ NN 
or LLM $|$ NN \\
or LLM \& Wiki/GPT-4 $|$ NN
\end{tabular}} 
& \multirow{2}{*}{
\begin{tabular}[c]{@{}c@{}}
LoRA
\end{tabular}} 
& \multirow{2}{*}{-} 
& \multirow{2}{*}{12} 
& Learning Rate & $\{10^{-6}, 10^{-5}, 10^{-4}, 10^{-3}, 0.01\}$  \\
& & & & Target adaptation Epoch & $\{25, 50, 100\}$ \\\midrule
\multirow{2}{*}{
\begin{tabular}[c]{@{}c@{}}
LLM \& Wiki/GPT-4 $|$ NN
\end{tabular}} 
& \multirow{2}{*}{
\begin{tabular}[c]{@{}c@{}}
Prefix Tuning
\end{tabular}} 
& \multirow{2}{*}{-} 
& \multirow{2}{*}{18} 
& Learning Rate & $\{10^{-5}, 10^{-4}, 10^{-3}, 0.01, 0.05, 0.1\}$  \\
& & & & Target adaptation Epoch & $\{25, 50, 100\}$ \\
\bottomrule
\end{tabular}}
\end{table}

\section{Additional Figures}
\label{sec: additional figs}

\paragraph{Shift Patterns} The shift patterns of all datasets are shown in~\Cref{fig-appendix:overall_decomposition}.

\begin{figure}[h]
\includegraphics[width=0.33\textwidth]{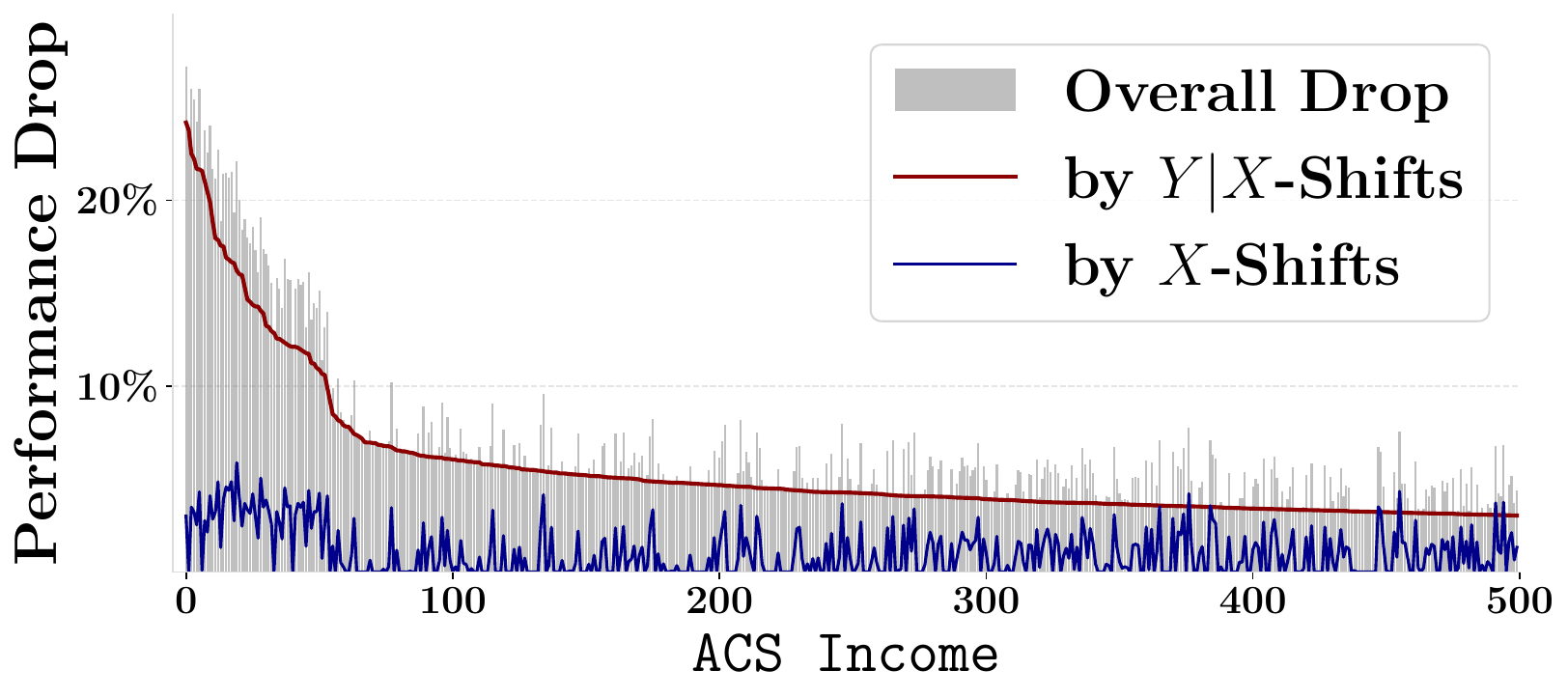}
\includegraphics[width=0.33\textwidth]{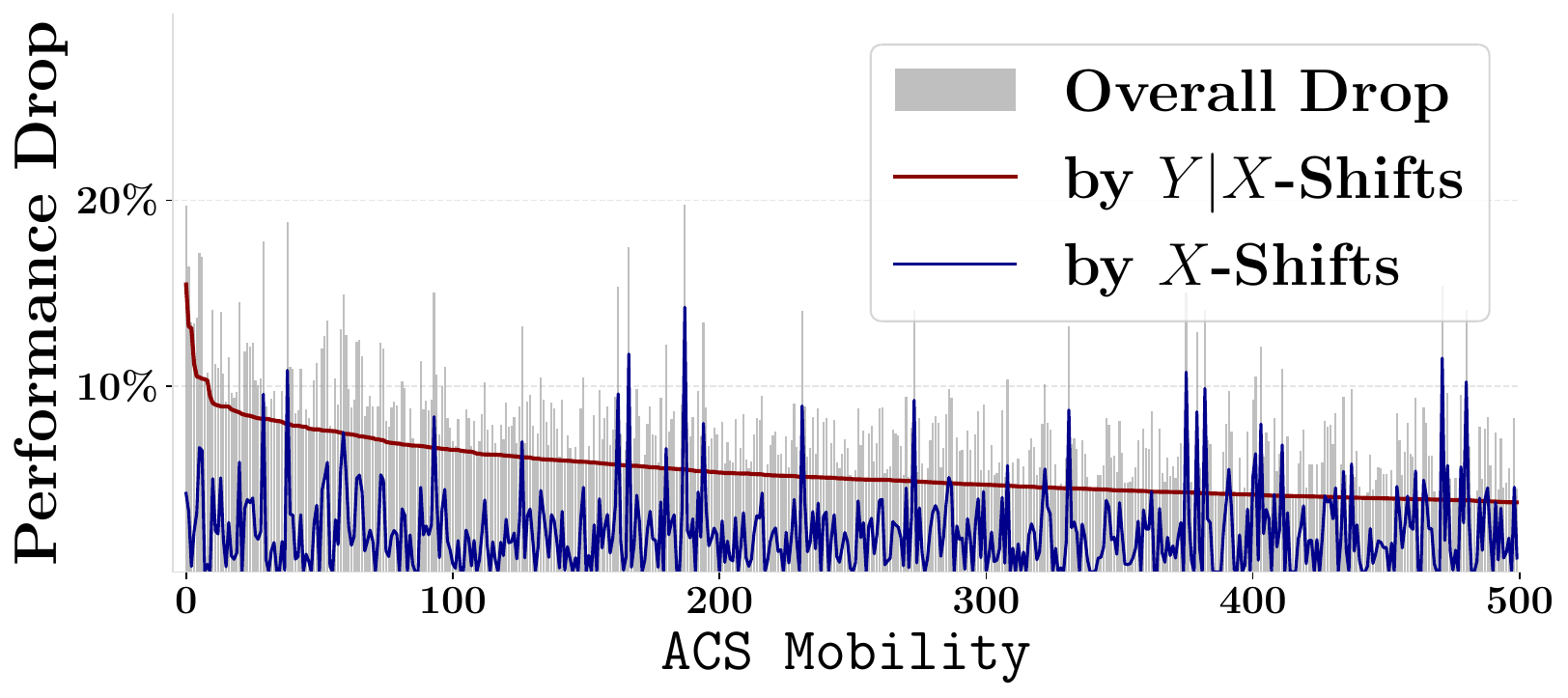}
\includegraphics[width=0.33\textwidth]{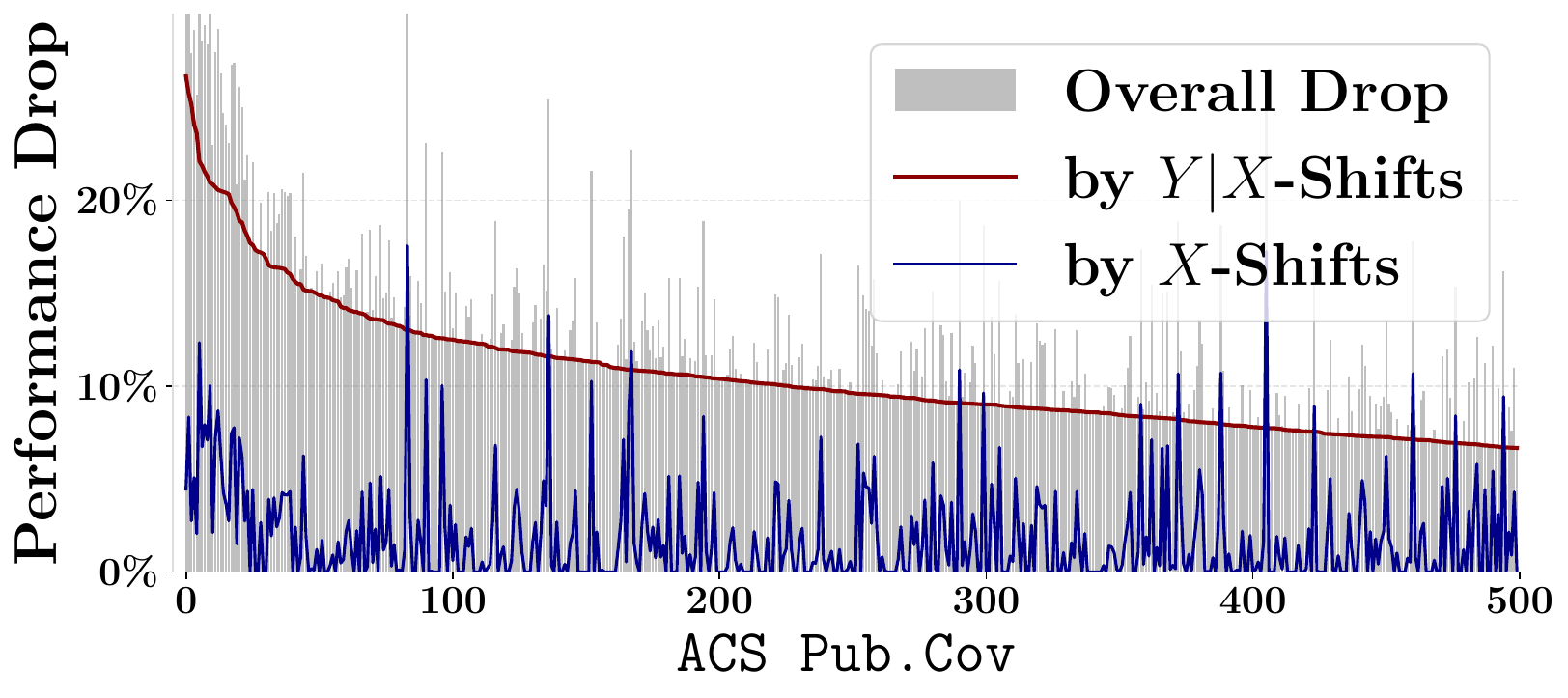}
\vspace{-1.5em}
\caption{Shift pattern analysis. For the 2550 source$\rightarrow$target distribution shift pairs in \texttt{ACS Income}, \texttt{ACS Pub.Cov}, \texttt{ACS Mobility} datasets respectively, we attribute the performance drop for each source$\rightarrow$target pair into $Y|X$-shifts (red curve) and $X$-shifts (blue curve), and sort all pairs according to the drop introduced by $Y|X$-shifts. We draw the \emph{worst-500} settings in each dataset, and the decomposition method used here is DISDE~\citep{CaiNaYa23} with XGBoost as the reference model.}
\label{fig-appendix:overall_decomposition}
\vspace{-1.5em}
\end{figure}

%\tableofcontents2
%\addtocontents{toc}{\protect\setcounter{tocdepth}{1}}  % This sets the depth of sections to be added to the ToC

%\addcontentsline{toc}{section}{Appendices} % Adds "Appendices" to the table of contents
%\section*{Appendices} % This provides a heading for the appendices but doesn't number it

\fi

%%%%%%%%%%%%%%%%%%%%%%%%%%%%%%%%%%%%%%%%%%%%%%%%%%%%%%%%%%%%%%%%%%%%%%%%%%%%%%%
%%%%%%%%%%%%%%%%%%%%%%%%%%%%%%%%%%%%%%%%%%%%%%%%%%%%%%%%%%%%%%%%%%%%%%%%%%%%%%%
% APPENDIX
%%%%%%%%%%%%%%%%%%%%%%%%%%%%%%%%%%%%%%%%%%%%%%%%%%%%%%%%%%%%%%%%%%%%%%%%%%%%%%%
%%%%%%%%%%%%%%%%%%%%%%%%%%%%%%%%%%%%%%%%%%%%%%%%%%%%%%%%%%%%%%%%%%%%%%%%%%%%%%%

\end{document}